\documentclass[10pt,journal,compsoc]{IEEEtran}

\usepackage{epsfig,graphicx,amsfonts,xcolor,multirow,amsmath,hyperref,booktabs,balance}
\usepackage[nocompress]{cite}
\usepackage{mycommand}
\usepackage{algorithm}
\usepackage{algorithmic}

\usepackage{amsmath,amsfonts,bm}

\def\eqref#1{equation~\ref{#1}}

\def\1{\bm{1}}

\def\mP{{\bm{P}}}

\def\mR{{\bm{R}}}

\DeclareMathAlphabet{\mathsfit}{\encodingdefault}{\sfdefault}{m}{sl}
\SetMathAlphabet{\mathsfit}{bold}{\encodingdefault}{\sfdefault}{bx}{n}

\usepackage{url}
\usepackage{array}
\usepackage{microtype}
\usepackage{graphicx}
\usepackage{xcolor}
\usepackage{subfigure}
\usepackage{amsmath}
\usepackage{bm}
\usepackage{dsfont}
\usepackage{mathtools}
\usepackage{nccmath}
\usepackage{amssymb}
\usepackage{multirow}
\usepackage{booktabs} 
\usepackage{varwidth}
\usepackage{pifont}
\usepackage{makecell}
\usepackage[font=small,labelfont=bf,tableposition=top]{caption}
\usepackage{wrapfig}
\usepackage[flushleft]{threeparttable}
\usepackage{varwidth}
\usepackage{pifont}
\usepackage{makecell}
\usepackage{enumitem}

\newcommand{\finntk}{\hat\Theta}

\newcommand{\Id}{\textbf{I}}

\begin{document}

\title{Understanding and Accelerating Neural Architecture Search with Training-Free and Theory-Grounded Metrics}

\author{
  Wuyang~Chen*,
  Xinyu~Gong*,
  Junru~Wu*,
  Yunchao~Wei,
  Humphrey~Shi,
  Zhicheng~Yan,\\
  Yi~Yang,
  and~Zhangyang~Wang

  \IEEEcompsocitemizethanks{
    \IEEEcompsocthanksitem Wuyang Chen, Xinyu Gong, and Zhangyang Wang are with the Department of Electrical and Computer Engineering, The University of Texas at Austin, TX, 78712. E-mail: \{wuyang.chen, xinyu.gong ,atlaswang\}@utexas.edu
    \IEEEcompsocthanksitem Junru Wu is with the Department of Computer Science, Texas A\&M University, TX, 77843. E-mail: sandboxmaster@tamu.edu
    \IEEEcompsocthanksitem Yunchao Wei is with the Institute of Information Science, Beijing Jiaotong University. E-mail: yunchao.wei@bjtu.edu.cn
    \IEEEcompsocthanksitem Humphrey Shi is with the Department of Computer and Information Science, University of Oregon. E-mail: hshi3@uoregon.edu
    \IEEEcompsocthanksitem Zhicheng Yan is with the Facebook AI Applied Research. E-mail: zyan3@fb.com
    \IEEEcompsocthanksitem Yi Yang is with Zhejiang University. E-mail: yangyics@zju.edu.cn
    \IEEEcompsocthanksitem The first two authors Wuyang Chen and Xinyu Gong contributed equally to this work. 
    \IEEEcompsocthanksitem  Correspondence to Zhangyang Wang (atlaswang@utexas.edu).
    \IEEEcompsocthanksitem The first three authors, Wuyang Chen, Xinyu Gong, and Junru Wu, contributed equally to this work. 
  }
}

\markboth{IEEE Transactions on Pattern Analysis and Machine Intelligence}
{Chen \MakeLowercase{\textit{et al.}}: TEG-NAS: Understanding and Accelerating Neural Architecture Search with Theory-Grounded Metrics}

\IEEEtitleabstractindextext{
  \begin{abstract}
    This work targets designing a principled and unified training-free framework for Neural Architecture Search (NAS), with high performance, low cost, and in-depth interpretation. NAS has been explosively studied to automate the discovery of top-performer neural networks, but suffers from heavy resource consumption and often incurs search bias due to truncated training or approximations. Recent NAS works \cite{mellor2021neural,chen2020tenas,abbdelfattah2020zero} start to explore indicators that can predict a network's performance without training. However, they either leveraged limited properties of deep networks, or the benefits of their training-free indicators are not applied to more extensive search methods.
    By rigorous correlation analysis, we present a unified framework to understand and accelerate NAS, by disentangling ``\textbf{TEG}'' characteristics of searched networks -- \textit{\textbf{T}rainability, \textbf{E}xpressivity, \textbf{G}eneralization} -- all assessed in a training-free manner. The TEG indicators could be scaled up and integrated with various NAS search methods, including both supernet and single-path NAS approaches. Extensive studies validate the effective and efficient guidance from our TEG-NAS framework, leading to both improved search accuracy and over 56\% reduction in search time cost. Moreover, we visualize search trajectories on three landscapes of ``\text{TEG}'' characteristics, observing that a good local minimum is easier to find on NAS-Bench-201 given its simple topology, whereas balancing ``\text{TEG}'' characteristics is much harder on the DARTS space due to its complex landscape geometry.
    Our code is available at \url{https://github.com/VITA-Group/TEGNAS}.
  \end{abstract}

}

\maketitle
\IEEEpeerreviewmaketitle

\IEEEraisesectionheading{
  \section{Introduction}
  \label{sec:introduction}
}

\IEEEPARstart{T}{he} development of deep convolutional neural networks significantly contributes to the success of computer vision tasks \cite{simonyan2014very,szegedy2015going,he2016deep,xie2017aggregated}. However, manually designing new network architectures not only costs tremendous time and resources, but also requires a rich network training experience that can hardly scale up.
Neural architecture search (\textbf{NAS}) is recently explored to remedy the human efforts and costs, benefiting automated discovery of architectures in a given search space \cite{zoph2016neural,brock2017smash,pham2018efficient,liu2018progressive,chen2018searching,bender2018understanding,gong2019autogan,fu2020autogan,chen2020fasterseg}.

Despite the principled automation, NAS still suffers from heavy consumption of computation time and resources. Most NAS methods mainly leverage the validation set and conduct accuracy-driven architecture optimization. Therefore, frequent training and evaluation of sampled architectures become a severe bottleneck that hinders both search efficiency and interpretation.
A super-network is extremely slow to be trained until converge \cite{liu2018darts} even with many effective heuristics for channel approximations or architecture sampling \cite{xu2019pc,dong2019searching}.
Approximated proxy inference such as truncated training/early stopping can accelerate the search, but is observed to introduce severe search bias \cite{pham2018efficient,liang2019darts+,tan2020efficientdet}.

People recently address this problem by proposing training-free NAS. Indicators like covariance of sample-wise Jacobian \cite{mellor2021neural}, Neural Tangent Kernel \cite{chen2020tenas}, and ``synflow'' \cite{abbdelfattah2020zero} are found to highly correlate with network's accuracy even at initialization (i.e., no gradient descent). This significantly reduces the search cost.
However, these works only validated a few \text{highly customized search approaches}, and leveraged limited properties of deep networks in an empirical or ad-hoc way. Mellor et al. \cite{mellor2021neural} only considered the ``local linear map'' defined by the covariance of sample-wise Jacobian, and only studied the random search method. Abdelfattah et al. \cite{abbdelfattah2020zero} mainly leveraged ``synflow'' proposed in previous pruning literature \cite{tanaka2020pruning} while relying on a warm-up stage. Chen et al. \cite{chen2020tenas} considered two aspects (trainability and expressivity) and integrated two indicators, but still have to leverage highly customized supernet pruning method and cannot extend to other non-supernet NAS search methods.
Moreover, these training-free indicators still only pursue final search performance and provide limited benefit towards the \text{interpretation and understanding} of the search trajectory and different search spaces.

In contrast, we target on designing a \textbf{unified} and \textbf{visualizable} training-free NAS framework that is (i) ``\textbf{search method agnostic}", i.e., can be scaled up to a broad variety of popular search algorithms; (ii)  ``\textbf{visualizable}", i.e., can help understand search behaviors on different landscapes of architecture spaces. Our core idea is to propose indicators that can rank network's performance, characterize network's properties, while still incurring no training cost. More importantly, we aim to make our training-free indicators widely 
applicable to multiple popular NAS methods, and also to facilitate the understanding of NAS search process.

Specifically, We first propose to disentangle the network's characteristics into three distinct aspects: \textit{\textbf{T}rainability, \textbf{E}xpressivity, \textbf{G}eneralization}, or ``\textbf{TEG}'' for short (defined in Section~\ref{sec:disentangling}). All three could be assessed with training-free indicators, and our studies demonstrate their strong correlations with network's training or test accuracy. Further, across various network operator types and topologies, they show complementary preferences, together leading to a comprehensive picture. 
Extensive studies validate the effective and efficient guidance from our TEG-NAS framework, with both improvements on search accuracy and over 56\% reduction on search time cost. More importantly, we for the first time visualize the search trajectory on architecture landscapes from different search spaces, thanks to our proposed TEG dimensions that disentangles different aspects of the searched model performance and can be efficiently quantified. For example, we find that a good local optimum is easier to find on NAS-Bench-201 \cite{dong2020bench} which has simpler topologies. However, it is much harder for a search method to balance TEG properties on the DARTS space with complex architecture landscapes.
We summarize our contributions as:
\begin{itemize}[leftmargin=*]
    \item We perform a rigorous correlation analysis of three disentangled ``\textbf{TEG}'' properties against network's training and test accuracy, and how changes made to an architecture will affect these aspects. All three notions are measured in a training-free manner. Since the three properties are complementary, they can achieve a very high correlation with the network's performance if properly combined.
    \item We design a unified training-free framework to provide accurate yet extremely efficient guidance during NAS search. Our framework is generally applicable to various existing NAS methods, including both supernet and single-path approaches, in a plug-and-play fashion. In both NAS-Bench-101, NAS-Bench-201, and DARTS search spaces, we trim down the search time by over 56\% while improving the searched model's accuracy.
    \item Beyond the final search performance, we for the first time visualize the search trajectory on the architecture landscapes from different search spaces, on how the search progresses along the TEG dimensions. That leads to a novel visualization of the NAS search process, as well as insightful comparison among different search spaces.
\end{itemize}

\textbf{Paper Organization.}
We first review recent advanced methods for efficient NAS and topics in Deep Learning theory in Section~\ref{sec:related_works}. We present our methods in two steps: 1) what are training-free indicators for NAS (Section~\ref{sec:disentangling}); 2) how to use training-free indicators in NAS (Section~\ref{sec:framework_section}).
In Section~\ref{sec:disentangling} we first introduce our motivation and background in analyzing trainability/expressivity/generalization of deep networks. Definitions and architecture inductive biases of three theory-grounded indicators will be explained, and we will demonstrate that disentangling different aspects of neural architectures leads to a better ranking prediction of networks from a search space.
After validating different preferences of our three training-free indicators on network architectures, in Section~\ref{sec:framework_section} we propose a unified and interpretable NAS framework that does not require any gradient descent training. Our NAS framework can not only be easily integrated into recent popular NAS methods (reinforcement learning, evolution, supernet), but also reflect a novel visualization of architecture landscapes. This contributes to both accelerated high-performance NAS methods and interpretable tools for analyzing NAS search space. We show our final results in Section~\ref{sec:exp}, where we studied the search accuracy and time cost on NAS-Bench-101~\cite{pmlr-v97-ying19a}, NAS-Bench-201~\cite{dong2020bench} and DARTS space.

The preliminary version of this work has been published in~\cite{chen2020tenas}, and we have made significant improvements over it. First, in the main method section, we will introduce a missing part in our ICLR version -- a training-free indicator for the generalization (Sec.~\ref{sec:generalization}). As generalization is a different property of deep networks besides trainability and expressivity, we will demonstrate its strong indication of network performance, its distinct architecture preference (Sec. ~\ref{sec:necessary_sufficient}), and its contribution to the final search results. Second, this version of training-free NAS is no longer a highly customized algorithm, but a unified and generally adoptable framework, which will be verified in three popular NAS search methods in our experiments (Sec. ~\ref{sec:exp}). All three NAS methods will benefit from strong search guidance and significant time cost reduction after being inter grated with our general framework. Finally, our new work will facilitate search space visualization and contribute to a novel visualization of the NAS search process. By tracking and projecting the search trajectory along the three proposed TEG dimensions, we can observe distinct landscape patterns from simple to complex search spaces, which will provide insights for understanding and designing NAS search spaces.

\section{Related Works}\label{sec:related_works}

\subsection{Neural Architecture Search}
Most NAS works suffer from heavy search costs.
Sampling-based methods \cite{pham2018efficient,real2019regularized,li2020gp,yang2020hournas,xu2021knas} achieve accurate network evaluations, but the truncated training imposes bias on the architecture rankings. The one-shot super network \cite{liu2018darts,kandasamy2018neural,dong2019searching,yu2020bignas,li2020sgas} can share parameters to sub-networks and greatly accelerate the evaluations, but it is hard to optimize \cite{yu2020train} and suffers from poor correlation between supernet accuracy and its sub-networks' \cite{yu2020evaluating}. In all, there is no clear one-winner method across the variety.

\subsection{Efficient and Training-free NAS}
Recent NAS works start focusing on reduced training or even training-free search. EcoNAS \cite{zhou2020econas} investigated different ad-hoc proxies (input size, model size, training samples, epochs, etc.) to reduce the training cost. Mellor et al. \cite{mellor2021neural} for the first time proposed a training-free NAS framework, which empirically leverages the correlation between sample-wise Jacobian to rank architectures. However, why did the Jacobian work was not clearly explained and demonstrated. Abdelfattah  et  al. \cite{abbdelfattah2020zero} studied different training-free indicators, and leveraged ``synflow'' from pruning \cite{tanaka2020pruning} as the main ranking indicator.
Park et al.~\cite{park2020towards} ranked the network's performance with NTK and NNGP.
Chen et al.~\cite{chen2020tenas} studied two theory-inspired indicators and combined with supernet pruning for further efficiency. However, these methods either leveraged ad-hoc or limited theory-driven properties of deep networks, or the benefits of their training-free strategies are tied to some specific search methods. In contrast, we hope to explore a comprehensive set of deep network properties, and further propose a unified training-free framework for various existing NAS methods.

\subsection{Trainability, Expressivity, and Generalization}
Numerous indicators in the deep learning theory field have been proposed to study various aspects of deep networks. Neural tangent kernel (NTK) is proposed to characterize the gradient descent training dynamics of wide networks \cite{jacot2018neural,hanin2019finite}. It was also proved that wide networks evolve as linear models under gradient descent \cite{lee2019wide}. Xiao et al. \cite{xiao2019disentangling} further propose to decouple the network's trainability and generalization. Meanwhile, a network's expressivity can be measured as the number of linear regions separated in the input space \cite{raghu2017expressive,serra2018bounding,hanin2019complexity,xiong2020number}. Many works also try directly probe network's generalization from various training statistics or network parameters \cite{jiang2019fantastic,lee2020neural,unterthiner2020predicting}.

\begin{figure*}[!t]
	\centering
	\includegraphics[width=0.35\textwidth]{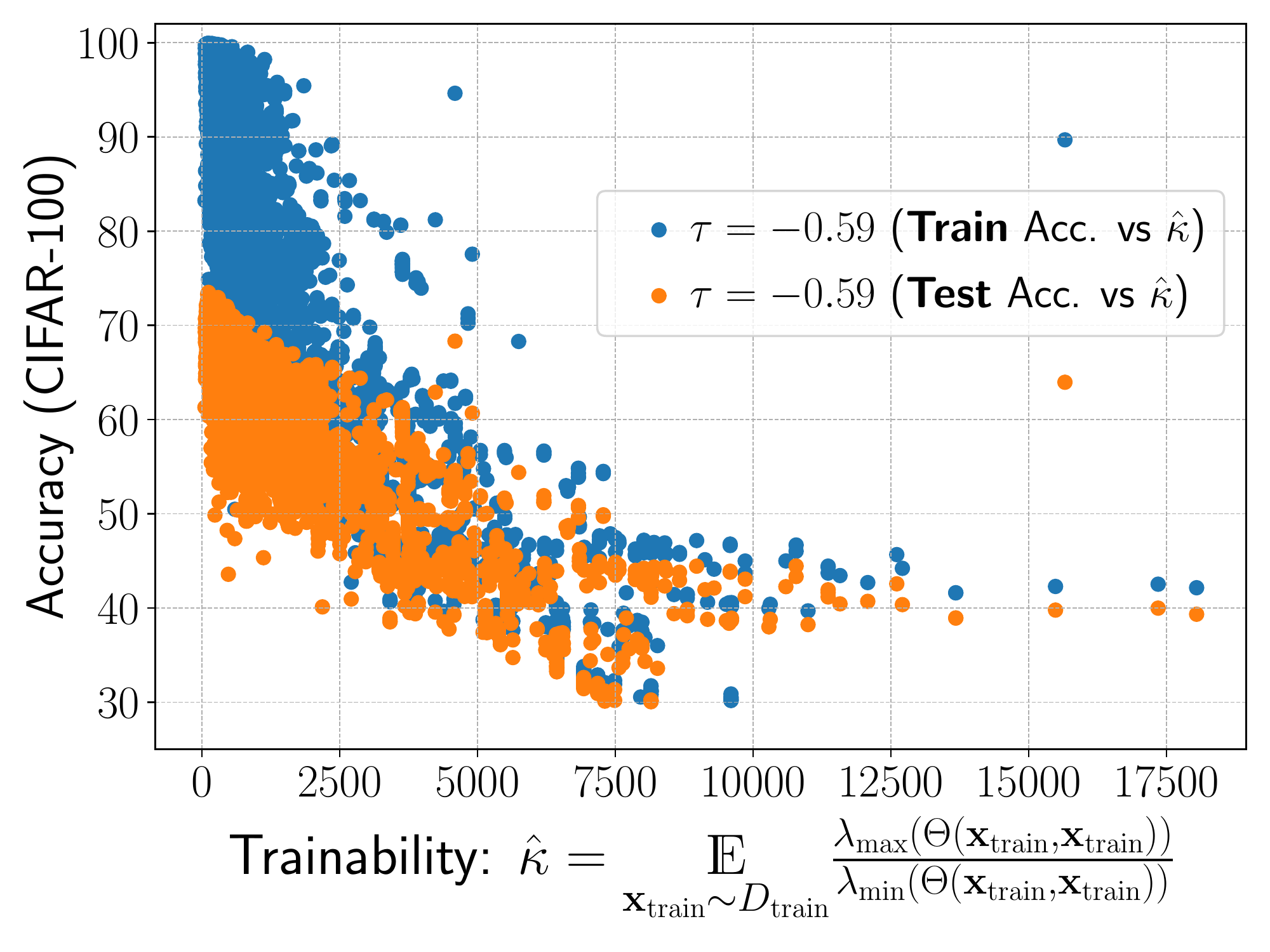}
	\includegraphics[width=0.35\textwidth]{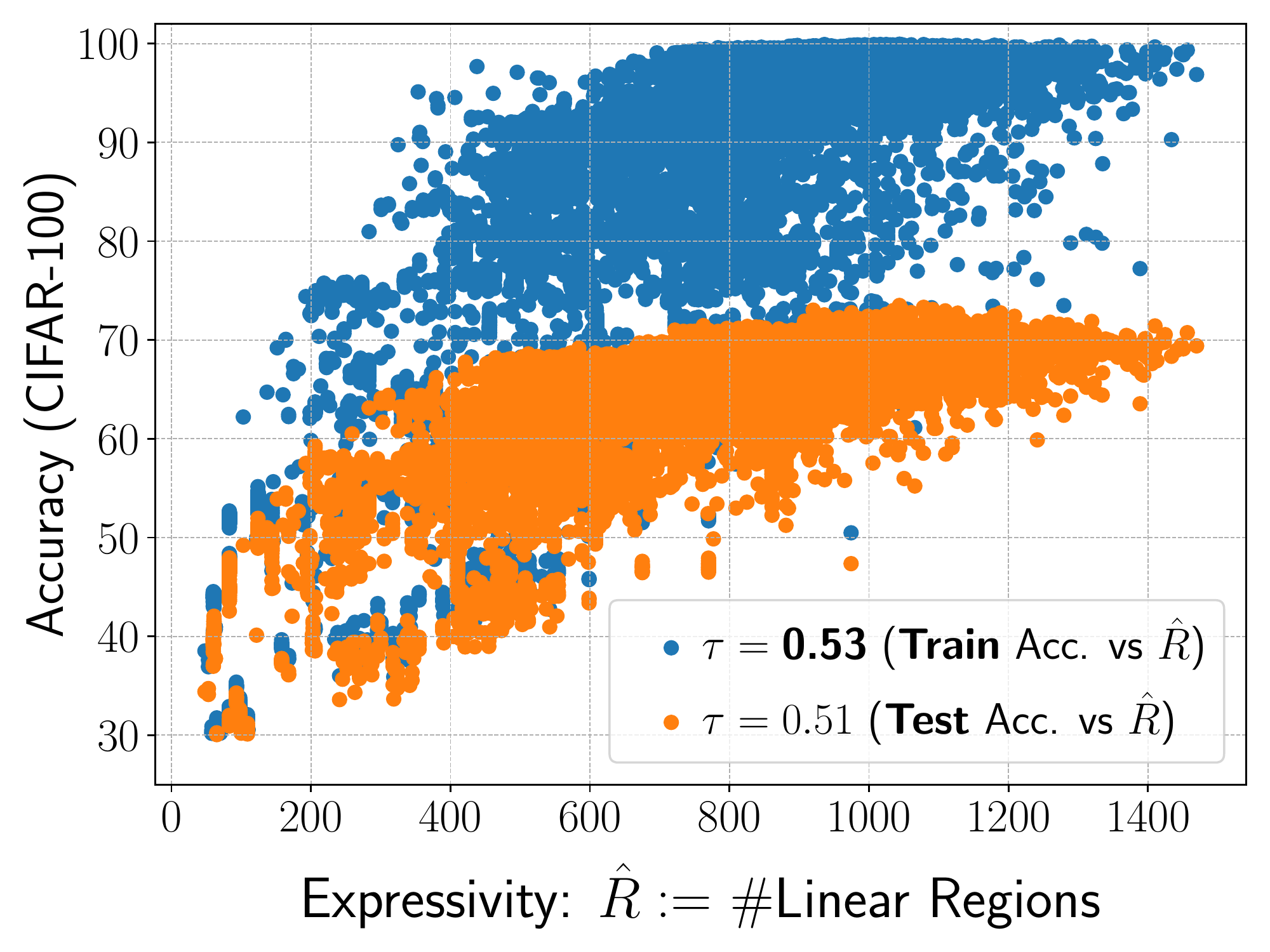}
	\includegraphics[width=0.35\textwidth]{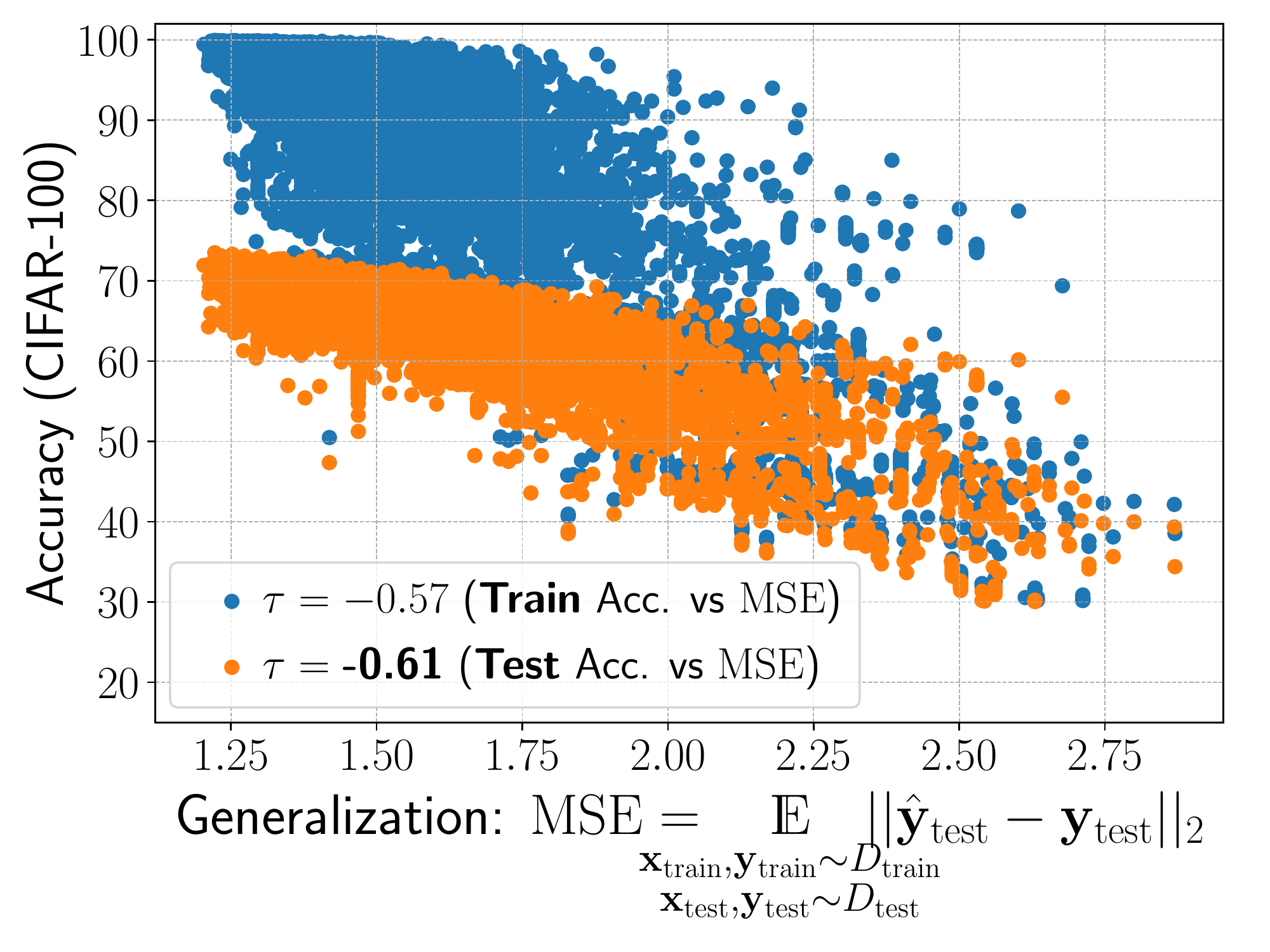}
	\includegraphics[width=0.35\textwidth]{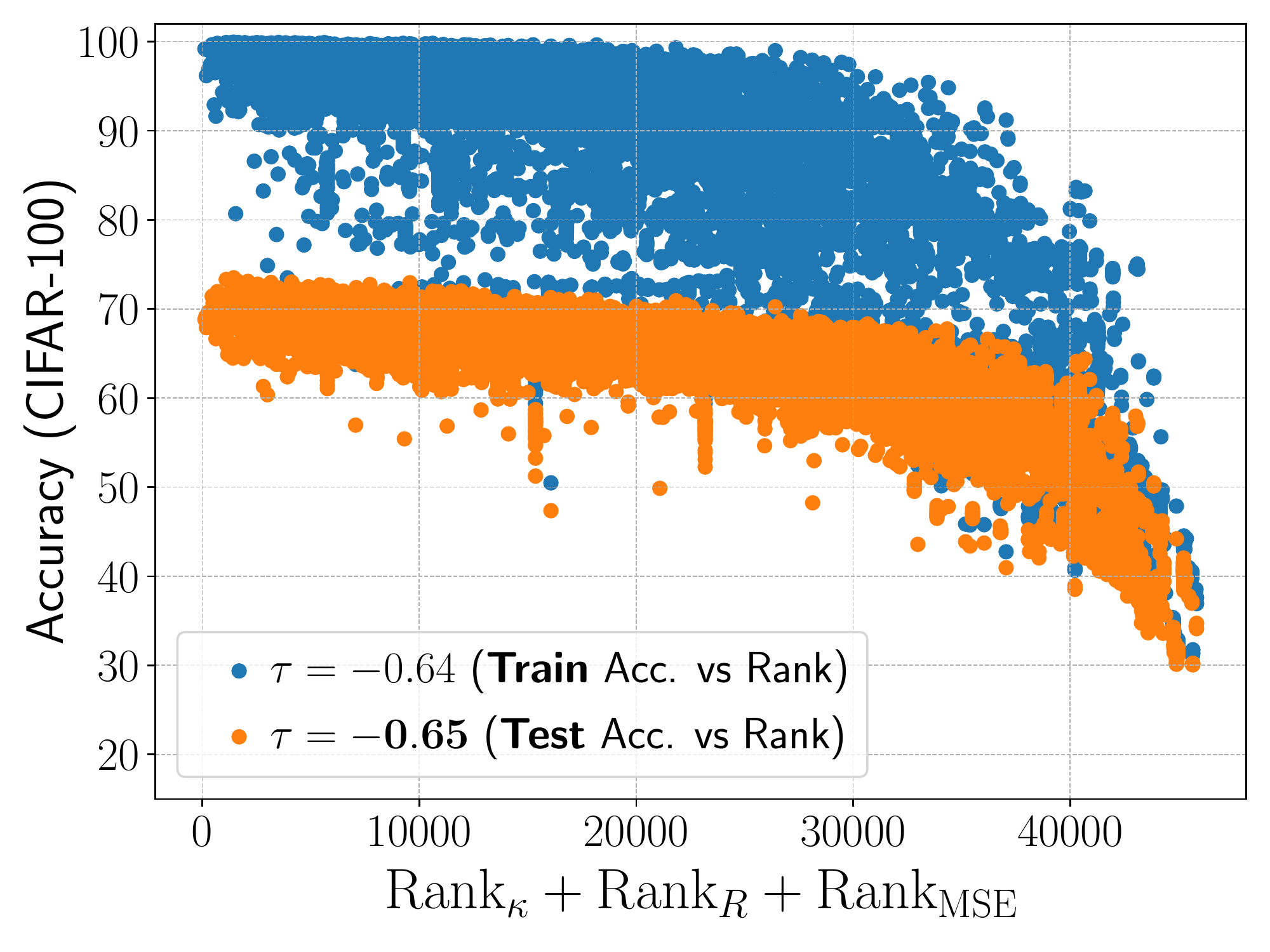}
	\caption{From left to right: correlation of trainability, expressivity, generalization, and sum of rankings against accuracies on NAS-Bench-201 \cite{dong2020bench}, all revealing strong correlations. Meanwhile, expressivity has stronger correlation against training accuracy, and generalization has stronger correlation against test accuracy, which are aligned with their definitions.}\label{fig:ntk_region_mse_201}
\end{figure*}

\section{Disentangling Trainability, Expressivity, and Generalization of Deep Networks} \label{sec:disentangling}

Trainability, expressivity, and generalization are three important, distinct, and complementary properties to characterize and understand neural networks \cite{hanin2019complexity,xiao2019disentangling,jiang2019fantastic}.
Specifically, the trainability is related to the convergence speed during optimization; the expressivity is related to the network's functional complexity; and the generalization indicates a model's error on unseen data.
Typically, a deep network achieves high performance when: 1) it can produce a loss landscape that is easily trainable with gradient descent, 2) it can represent sufficiently complex functions, 3) it can learn representation transferable to unseen examples, instead of just memorizing training data.
In this section, we will introduce what are these training-free indicators, and in Section~\ref{sec:framework_section} we will introduce how to use them in NAS.

\subsection{Trainability}

Training deep networks requires optimizing high-dimensional non-convex loss functions.
In practice, gradient descent often finds the global or good local minimum. However, many expressible networks are not easily learnable. For example, a deep stack of convolutional layers (e.g. Vgg \cite{simonyan2014very}) is much harder to train than networks with skip connections (ResNet \cite{he2016deep}, DenseNet \cite{huang2017densely}, etc.), even the former could equip a larger number of parameters.
The trainability of a neural network studies how effective it can be optimized by gradient descent \cite{burkholz2019initialization,hayou2019impact,shin2020trainability}.

\textbf{Architecture Bias on Trainability:} 
A network's architecture can control how effectively the gradient information can flow through it. These topological properties might control the amount of information that can be learned by networks.
Preserving the gradient flow is found to be essential during network pruning, even at initialization \cite{wang2020picking,tanaka2020pruning}. Skip connections also have a significant impact on the sharpness/flatness of the loss landscapes \cite{li2017visualizing}.
Therefore, we hypothesize that certain aspects of trainability can be characterized just by the architecture at its initialization.

\textbf{Conditioning of NTK:} To characterize the training dynamics of wide networks, Neural tangent kernel (NTK) is proposed \cite{Jacot2018ntk, lee2019wide, chizat2019lazy}, defined as:
{\normalsize
  \setlength{\abovedisplayskip}{2pt}
  \setlength{\belowdisplayskip}{\abovedisplayskip}
  \setlength{\abovedisplayshortskip}{10pt}
  \setlength{\belowdisplayshortskip}{2pt}
  \begin{equation}
    \finntk(\bm{x}, \bm{x}') = J(\bm{x})J(\bm{x}')^T,
  \end{equation}
}where $J(\bm{x})$ is the Jacobian evaluated at a point $\bm{x}$.
Xiao et al. \cite{xiao2019disentangling} measures the trainability of networks by studying the spectrum and conditioning of $\finntk$:
{\normalsize
  \setlength{\abovedisplayskip}{2pt}
  \setlength{\belowdisplayskip}{2pt}
  \setlength{\abovedisplayshortskip}{2pt}
  \setlength{\belowdisplayshortskip}{5pt}
  \begin{align}
    \mu_t(\bm{x}_{\text{train}}) &= (\Id - e^{-\eta\finntk(\bm{x}_\mathrm{train},\bm{x}_\mathrm{train}) t})\bm{y}_\mathrm{train}\label{eq:fc_ntk_recap_dynamics}\\
    \mu_t(\bm{x}_{\text{train}})_i &= (\Id - e^{-\eta \lambda_i t}) \bm{y}_{{\text{train}},i}\label{eq:fc_ntk_dynamics_eigen}.
  \end{align}
}$\mu_t(\bm{x})$
is the expected outputs of a wide network, $\lambda_i$ are the eigenvalues of $\finntk(\bm{x}_\mathrm{train},\bm{x}_\mathrm{train})$,
$\bm{x}_{\text{train}}$ and $\bm{y}_{\text{train}}$ are drawn from the training set $D_{\text{train}}$.
Eq. \ref{eq:fc_ntk_dynamics_eigen} indicates that different time is needed to learn the $i$-th eigenmode, and thus we can conclude that the more diverse the learning speeds of different eigenmodes are, the more difficult the network can be optimized. We are therefore motivated to use the empirical condition number of NTK to represent trainability:
{\normalsize
  \setlength{\abovedisplayskip}{3pt}
  \setlength{\belowdisplayskip}{\abovedisplayskip}
  \setlength{\abovedisplayshortskip}{0pt}
  \setlength{\belowdisplayshortskip}{3pt}
\begin{equation}
    \hat{\kappa} = \underset{\substack{\bm{x}_\mathrm{train} \sim D_{\mathrm{train}} \\ \theta \sim \mathcal{N}(0, \frac{2}{N_l})}}{\mathds{E}} \frac{\lambda_\mathrm{max}(\finntk(\bm{x}_\mathrm{train},\bm{x}_\mathrm{train}))}{\lambda_\mathrm{min}(\finntk(\bm{x}_\mathrm{train},\bm{x}_\mathrm{train}))},
\end{equation}
}where network parameters $\theta$ are drawn from Kaiming normal initialization $\mathcal{N}(0, \frac{2}{N_l})$ ($N_l$ is the width at layer $l$) \cite{he2015delving}, and thus $\hat{\kappa}$ is calculated at network's initialization.
As shown in Figure \ref{fig:ntk_region_mse_201}, $\hat{\kappa}$ is negatively correlated with both the network's training and test accuracy, with the Kendall-tau correlation as $-0.59$.
Therefore, minimizing the $\hat{\kappa}$ during the search will encourage the discovery of architectures with high performance.

\subsection{Expressivity}

Recent works try to explain the success of deep networks by their ability to approximate complex functions, quantified by various complexity measures \cite{cohen2016expressive,croce2019provable}. The more expressible the network is, the more efficient it can fit the training data.
In the case of ReLU networks that compute piecewise linear functions, the number of distinct linear regions is a natural measure of such expressivity.
The composition of ReLU leads the input space partitioned into distinct pieces (i.e. linear regions).
Therefore, the density of linear regions serves as a convenient proxy for the complexity of the network \cite{raghu2017expressive,hanin2019complexity,hanin2019deep,xiong2020number}.

\textbf{Architecture Bias on Expressivity:}
It was proved that networks with random Gaussian initialization can embed the training data in a distance-preserving manner \cite{giryes2016deep}.
Hanin et al. \cite{hanin2019deep} show that the number of activation patterns for ReLU networks is tightly bounded by the total number of neurons both at initialization and during training, and empirically showed that the number of regions stays roughly constant during training \cite{hanin2019complexity}.
Therefore, network architecture itself has a strong inductive bias on its expressivity.

\textbf{Complexity of Linear Regions:}
We first introduce the definition of activation patterns and see how it is connected to the number of linear regions in input space.

\textbf{Definition 1 of \cite{xiong2020number} (Activation Patterns as the Linear Regions)}\label{def:activation-regions}
\textit{
Let $\mathcal{N}$ be a ReLU CNN. 
An activation pattern of $\mathcal{N}$ is a function $\mP$ from the set of neurons to $\{1,-1\}$, i.e., for each neuron $z$ in $\mathcal{N}$, we have $\mP(z) \in \{1,-1\}$.
Let $\theta$ be a fixed set of parameters (weights and biases) in $\mathcal{N}$, and $\mP$ be an activation pattern. The region corresponding to $\mP$ and $\theta$ is
\begin{multline}
    \mR(\mP;\theta) := \{\bm{x}^0\in \mathbb{R}^{C\times H \times W}: z(\bm{x}^0;\theta)\cdot {\mP(z)} >0, \\ \quad \forall z \in \mathcal{N}\},
    \label{eq:linear_region}
\end{multline}
where $z(\bm{x}^0;\theta)$ is the pre-activation of a neuron $z$.
Let $R_{\mathcal{N},\theta}$ denote the number of linear regions of $\mathcal{N}$ at $\theta$, i.e., $R_{\mathcal{N},\theta} := \#\{  \mR(\mP;\theta): \mR(\mP;\theta)\neq \emptyset ~ \text{ for some activation pattern } \mP    \}.$
}

Eq. \ref{eq:linear_region} tells us that a linear region in the input space is a set of input data $\bm{x}^0$ that satisfies a certain fixed activation pattern $\mP(z)$, and therefore the number of linear regions $R_{\mathcal{N},\theta}$ measures how many unique activation patterns that can be divided by the network.

Since the input space is recursively partitioned by ReLU as the layers go deeper,
and the composition of piecewise linear functions is still piecewise linear,
each linear region in the input space can be uniquely represented with a set of affine parameters based on a combination of ReLU activation patterns. This means that, with given training examples and parameters, the number of linear regions $R(\bm{x}_\mathrm{train}, \theta)$ can be approximated by the number of unique activation patterns combined from all ReLU layers in the whole network.
We are therefore motivated to use the empirical number of linear regions to represent expressivity:
{\normalsize
  \setlength{\abovedisplayskip}{0pt}
  \setlength{\belowdisplayskip}{\abovedisplayskip}
  \setlength{\abovedisplayshortskip}{-5pt}
  \setlength{\belowdisplayshortskip}{2pt}
\begin{equation}
    \hat{R} = \underset{\substack{\bm{x}_\mathrm{train} \sim D_{\mathrm{train}} \\ \theta \sim \mathcal{N}(0, \frac{2}{N_l})}}{\mathds{E}} R(\bm{x}_\mathrm{train}, \theta).
\end{equation}
}As shown in Figure \ref{fig:ntk_region_mse_201}, $\hat{R}$ is positively correlated with both the network's training and test accuracy. Moreover, we observe that $\hat{R}$ has a stronger correlation with training over test accuracy, which validates that $\hat{R}$ indicates how well a network fits the training data, but not its generalizability.

\subsection{Generalization} \label{sec:generalization}
Typically, the generalization error\footnote{Here we quantify the absolute generalization error, instead of the generalization gap.} is defined as the risk of the model over the underlying data distribution $D$. A model chosen from a very complex family of functions can essentially fit all the training data, but memorization cannot guarantee the accurate association of unseen examples with seen ones. That makes generalization a distinct notion from trainability (``optimization'') and expressivity (``memorization''), since generalization focuses on how well a model can transfer the information from seen to unseen data.

\textbf{Architecture Bias on Generalization:} With even random initialization, network architecture alone could have a strong inductive bias to its generalization error.
Network architectures of different complexity or sparsity, without learning any weight parameters, are found to be able to encode solutions for a given task \cite{gaier2019weight,ramanujan2020s}.
Bhardwaj et al. \cite{bhardwaj2019towards} formally established a link between the structure of CNN architectures (depths, widths, number of skip connections, etc.) and their generalization errors.
More importantly, inductive bias from certain architecture patterns (e.g. graph-based representation \cite{you2020graph}) can even transfer across different types of networks (MLPs, CNNs, ResNets, etc.) and different tasks (CIFAR-10, ImageNet, etc.).
Same in our work, we decouple the architecture from the network weights, and focus only on the aspect of “weight-agnostic” generalization, which is impacted by just the network architecture.

\textbf{NTK Kernel Regression:}
Previous works \cite{lee2019wide,xiao2019disentangling} showed that at time $t$ during gradient descent training with an MSE loss, the expected outputs of an infinite wide network evolve as:
{\normalsize
  \setlength{\abovedisplayskip}{2pt}
  \setlength{\belowdisplayskip}{5pt}
  \setlength{\abovedisplayshortskip}{2pt}
  \setlength{\belowdisplayshortskip}{5pt}
\begin{multline}
    \mu_{t}(\bm{x}_\mathrm{test}) = \finntk(\bm{x}_\mathrm{test},\bm{x}_\mathrm{train})(\finntk(\bm{x}_\mathrm{train},\bm{x}_\mathrm{train}))^{-1}(\bm{I} -\\
    e^{-\eta \finntk(\bm{x}_\mathrm{train},\bm{x}_\mathrm{train})t})\bm{y}_\mathrm{train}.
    \label{eq:mu_t_wide}
\end{multline}
}Studying the evolution of $\mu_{t}(\bm{x}_\mathrm{test})$ in Eq. \ref{eq:mu_t_wide} along the training iteration $t$ can characterize the generalization performance of deep networks. However, the infinite width is not directly applicable in real-life scenarios, and we want to estimate the generalization at a network's initialization. Therefore in our work, we choose to empirically estimate the generalization by calculating the test MSE error of a network's NTK kernel regression:
{\setlength{\abovedisplayskip}{2pt}
  \setlength{\belowdisplayskip}{-5pt}
  \setlength{\abovedisplayshortskip}{2pt}
  \setlength{\belowdisplayshortskip}{-5pt}
\begin{equation}\begin{medsize}
    \hat{\bm{y}}_\mathrm{test} = \finntk^L(\bm{x}_\mathrm{test},\bm{x}_\mathrm{train})(\finntk^L(\bm{x}_\mathrm{train},\bm{x}_\mathrm{train}))^{-1}\bm{y}_\mathrm{train},\label{eq:ntk_y_hat}
\end{medsize}\end{equation}
}{\setlength{\abovedisplayskip}{0pt}
  \setlength{\belowdisplayskip}{1pt}
  \setlength{\abovedisplayshortskip}{0pt}
  \setlength{\belowdisplayshortskip}{1pt}
\begin{equation}\begin{medsize}
    \mathrm{MSE} = \underset{\substack{\bm{x}_\mathrm{train}, \bm{y}_\mathrm{train} \sim D_{\mathrm{train}} \\ \bm{x}_\mathrm{test}, \bm{y}_\mathrm{test} \sim D_{\mathrm{test}}}}{\mathds{E}} ||\hat{\bm{y}}_\mathrm{test} - \bm{y}_\mathrm{test}||_2.\label{eq:ntk_mse}
\end{medsize}\end{equation}
}$\finntk^L$ indicates the NTK evaluated only for the last layer of the deep network. Eq. \ref{eq:ntk_y_hat} tries to associate $\bm{x}_\mathrm{test}$ with $\bm{x}_\mathrm{train}$ via NTK kernel regression, and transfer the given training labels $\bm{y}_\mathrm{train}$ to the test data.
A deep neural network will fail to generalize if its prediction $\hat{\bm{y}}_\mathrm{test}$ becomes data-independent, and the $\mathrm{MSE}$ in Eq. \ref{eq:ntk_mse} will become large.

Note that we are not directly predicting a network's converged generalization at its initialization. Instead, we use $\mathrm{MSE}$ to compare different networks and study how it would be affected by different architectures. Adopting $\mathrm{MSE}$ also follows the convention that NTK is also derived under the squared loss \cite{jacot2018neural,arora2019exact}.

As further demonstrated in Figure \ref{fig:ntk_region_mse_201}, $\mathrm{MSE}$ shows strong negative correlation with both the network's training accuracy and test accuracy.
We also observe that both training and testing accuracy drops with the increasing of $\mathrm{MSE}$. This is because on the observation from NAS-Bench-201, that all models' training and testing accuracies are positively correlated.
More importantly, $\mathrm{MSE}$ has a stronger correlation with the test than the training accuracy. This precisely validates that $\mathrm{MSE}$ represents how well a network generalizes, but not memorization of the training data.

At this moment we disentangled the network's performance into three distinct properties. In the next section, we present our unified and interpretable TEG-NAS strategy.

\section{TEG-NAS: a Unified and Interpretable NAS Framework} \label{sec:framework_section}
In this section we will demonstrate how to use our three training-free indicators in NAS.
Our core motivation is to provide a unified training-free framework for NAS of both high performance and low cost.
We also enable the visualization of NAS search trajectory on the architecture landscapes.

\subsection{How Architecture Affects $\hat{\kappa}$, $\hat{R}$, and $MSE$}\label{sec:necessary_sufficient}

Despite the strong correlations and different preferences over training or testing accuracy we observe in Figure \ref{fig:ntk_region_mse_201}, it is still unknown whether each individual aspect of three -- trainability, expressivity, generalization -- is necessary for a deep network to be of high performance. This analysis is also missing in previous works \cite{xiao2019disentangling,chen2020tenas}.
Before we directly adopt our disentangled \textbf{TEG} properties to NAS search, we must study how changes of $\hat{\kappa}$, $\hat{R}$, $\mathrm{MSE}$ could be reflected on network architectures, and how network's operator types or topology will affect its trainability, expressivity, and generalization. Otherwise, if they share the same preference on selecting architectures, picking any one of them will guide the search towards similar results.

\textbf{Architecture Exclusively Selected by $\hat{\kappa}$, $\hat{R}$, $\mathrm{MSE}$}

\begin{figure}[!h]
\includegraphics[scale=0.5]{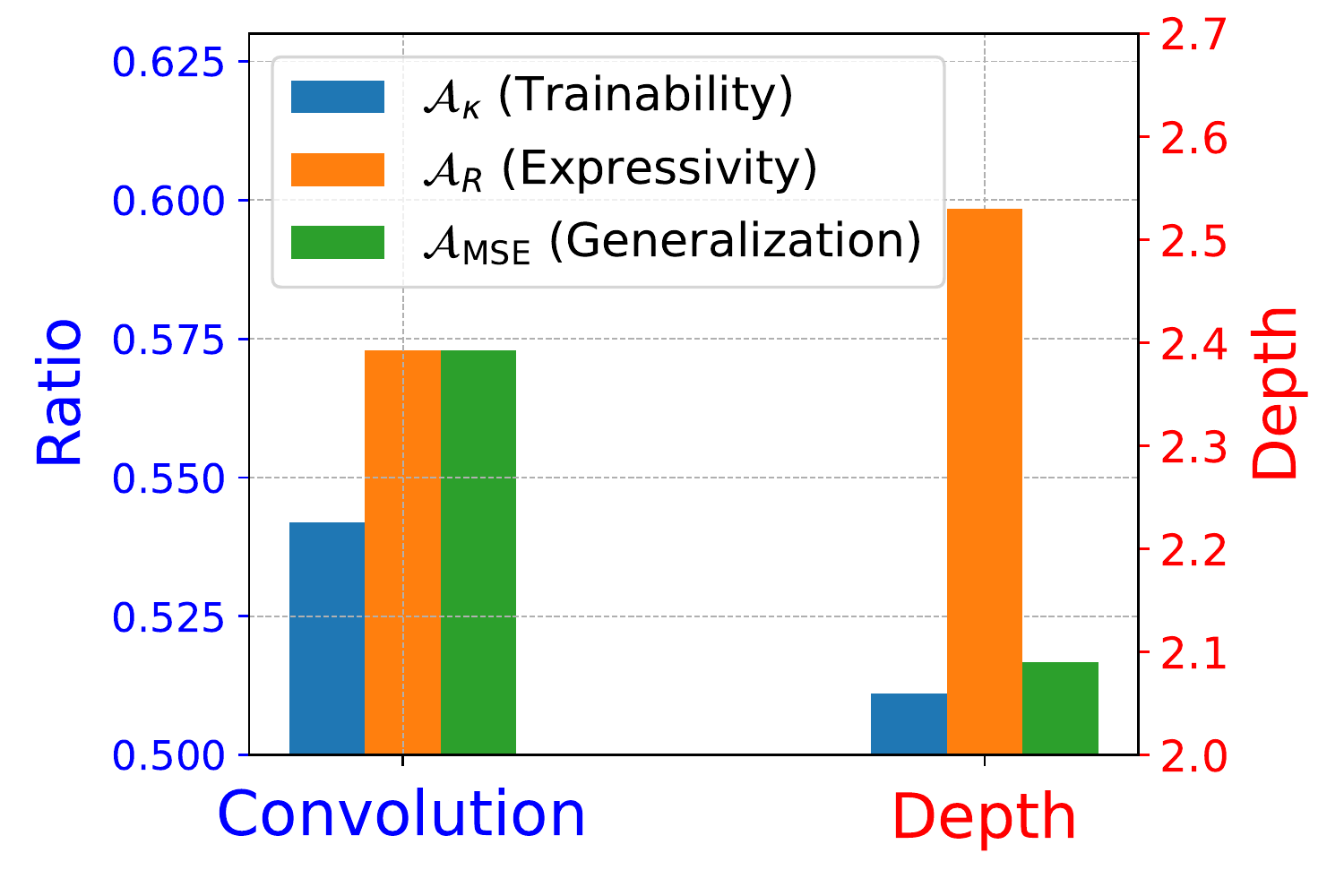}
\centering
\caption{Impact of network operators and topologies on $\hat{\kappa}$, $\hat{R}$, and $MSE$ on NAS-Bench-201.}
\label{fig:op_preference}
\end{figure}

Trainability, expressivity, and generalization may have different preferences over network's operator types and topology. This motivates us to summarize architectures that are exclusively selected by $\hat{\kappa}$, $\hat{R}$, $\mathrm{MSE}$.
We first measure the thresholds $T_{\kappa}$, $T_{R}$, $T_{\mathrm{MSE}}$ that filter top 10\% architectures out of the search space $\bm{A}$, ranked by $\hat{\kappa}$, $\hat{R}$, and $\mathrm{MSE}$, respectively. We define the following three subsets of architectures,
with any two out of three having an empty intersection:
{\normalsize
  \setlength{\abovedisplayskip}{2pt}
  \setlength{\belowdisplayskip}{2pt}
  \setlength{\abovedisplayshortskip}{2pt}
  \setlength{\belowdisplayshortskip}{2pt}
\begin{align}
    &\medmath{\bm{A}_{\kappa} = \{a | a \in \bm{A}, \hat{\kappa}_a \leq T_{\kappa}, \hat{R}_a < T_{R}, \mathrm{MSE}_a > T_{\mathrm{MSE}} \}},\\
    &\medmath{\bm{A}_{R} = \{a | a \in \bm{A}, \hat{\kappa}_a > T_{\kappa}, \hat{R}_a \geq T_{R}, \mathrm{MSE}_a > T_{\mathrm{MSE}} \}},\\
    &\medmath{\bm{A}_{{\tiny \mathrm{MSE}}} = \{a | a \in \bm{A}, \hat{\kappa}_a > T_{\kappa}, \hat{R}_a < T_{R}, \mathrm{MSE}_a \leq T_{\mathrm{MSE}} \}}.
\end{align}}
We study three subsets of architectures in terms of both operator and topology, shown in Figure \ref{fig:op_preference}. For operator types, the ratio of convolution ($1\times 1$ and $3\times 3$) operators in $\bm{A}_{\kappa}$ is lower than those of $\bm{A}_{R}$ and $\bm{A}_{\mathrm{MSE}}$, indicating operators with a heavy number of parameters may not friendly for optimization. In contrast, $\bm{A}_{R}$ and $\bm{A}_{\mathrm{MSE}}$ favor more convolution layers, benefiting to data fitting. For network topology, the averaged depth\footnote{Depth of a cell is defined as the number of connections on the longest path from input to the output \cite{shu2019understanding}} of architectures in $\bm{A}_{\kappa}$ is much lower than those in $\bm{A}_{R}$, since shallow networks are easier to train \cite{xiao2019disentangling}. Depth from $\bm{A}_{\mathrm{MSE}}$ is also low, contributing to better test accuracy.

\begin{figure}[h!]
\includegraphics[scale=0.36]{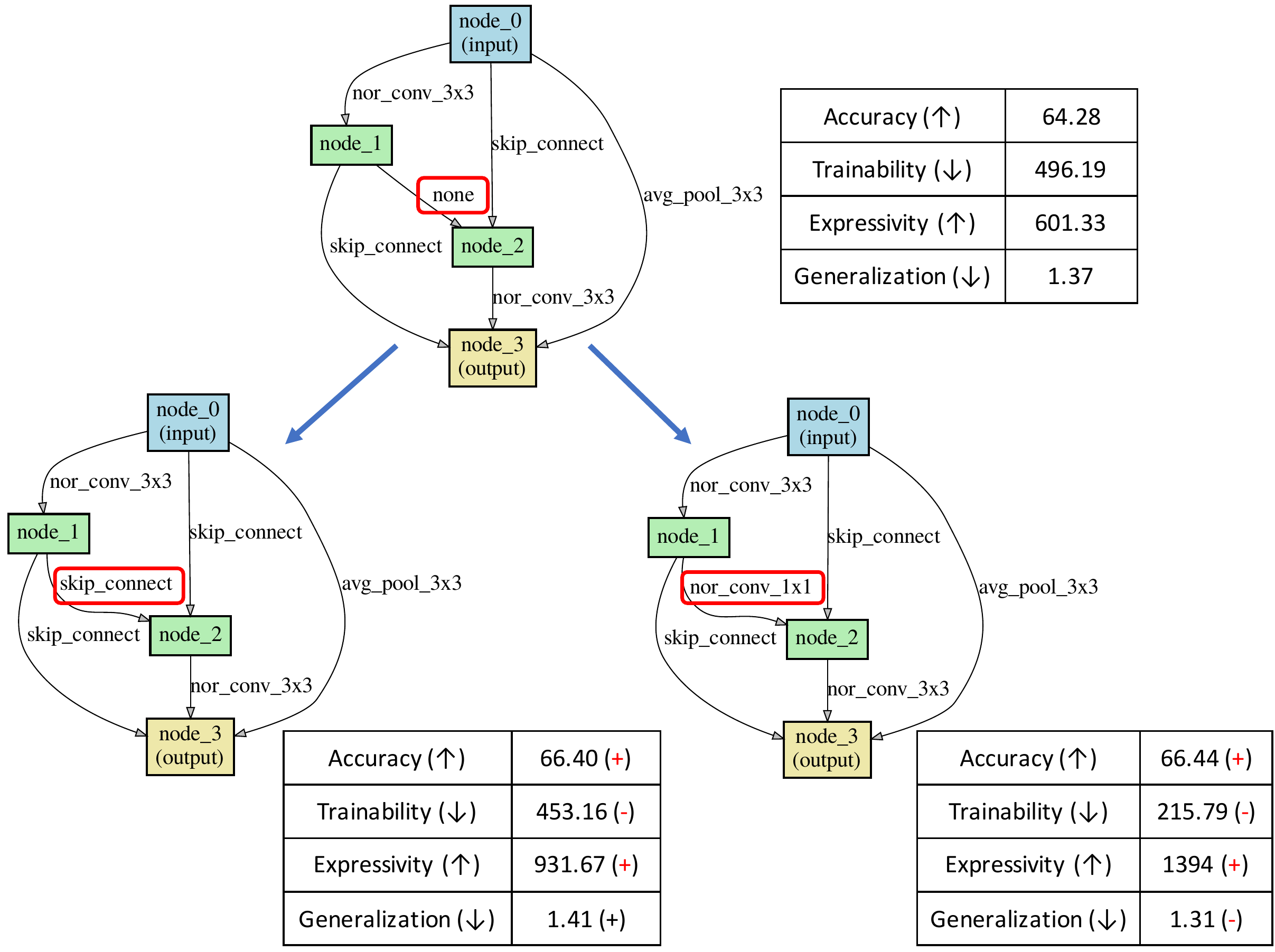}
\centering
\caption{``None'' operator jeopardize the architecture's trainability, expressivity, and generalization. By replacing ``none'' with ``skip\_connect'' or ``conv$1\times 1$'', the bad trainability or expressivity can be addressed, leading to better test accuracy.}
\label{fig:show_case}
\end{figure}

\begin{table*}[!t]
    \centering
    \caption{Comparison of different NAS search methods studied in our experiments.}
    \resizebox{0.92\textwidth}{!}{
    \begin{tabular}{ccccc}
        \toprule
        NAS Method & NAS Formulation & Weight-sharing & Update Method & Search Stopping Criterion \\ \midrule
        REINFORCE & Single-path & No & policy gradients & policy entropy \\\midrule
        Evolution & Single-path & No & mutation & population diversity \\\midrule
        FP-NAS & Supernet & Yes & gradient descent & entropy of architecture parameters \\
        \bottomrule
    \end{tabular}\label{table:nas_method_comparison}
    }
\end{table*}

\textbf{Case Study}

One failure reason for a bad architecture is the exist of ``none'' (or ``zero'') operator, which completely breaks the feed-forward and gradient flow. We show one case in Figure \ref{fig:show_case}. When there is a ``none'' exist, both trainability and expressivity are bad, leading to poor test accuracy. By switching into ``skip\_connect'' or ``conv$1\times 1$'', the bad trainability or expressivity are addressed, leading to better test accuracy.

\subsection{A Unified Training-free NAS Framework} \label{sec:framework}

Different preferences of $\hat{\kappa}$, $\hat{R}$, and $\mathrm{MSE}$ on network's operators and topology validate their potential of guiding the NAS search. We now propose our unified training-free NAS framework (Algorithm \ref{algo:framework}). Existing NAS methods evaluate the accuracy or loss value of every single architecture via truncated training or shared supernet weights. The evaluated accuracy or loss is also leveraged as feedback to update the NAS method itself. Instead, we leverage the disentangled trainability, expressivity, and generalization during the search. For each architecture sampled by the NAS search method, we average three repeated calculations of $\hat{\kappa}_t$ / $\hat{R}_t$ / $\mathrm{MSE}_t$, by using three independent mini-batches of training data. They will be leveraged as the feedback to guide the update of the search method.
For different search stopping criteria and update manners of different NAS methods, please refer to Section \ref{sec:nas_method_comparison} and Table \ref{table:nas_method_comparison}.
\begin{algorithm}[hb!]
\footnotesize
	\begin{algorithmic}[1]
	    \STATE {\bfseries Input:} architecture search space $\bm{A}$, NAS search method $\mathcal{M}$, step $t = 0$.
    	\WHILE{\textbf{not} Search Stopping Criterion of $\mathcal{M}$ satisfied}
    	    \STATE Sample architecture: $a_t$ = $\mathcal{M}$.sample($\bm{A}$)
    	    \STATE Calculate $\hat{\kappa}_t$, $\hat{R}_t$, $\mathrm{MSE}_t$ for $a_t$
    	    \STATE Update NAS method: $\mathcal{M}$.update($a_t$, $\hat{\kappa}_t$, $\hat{R}_t$, $\mathrm{MSE}_t$)
    	    \STATE $t = t + 1$
    	\ENDWHILE
    	\STATE {\bfseries Return} Searched architecture $\mathcal{M}$.derive().
	\end{algorithmic}
	\caption{Our unified training-free framework for different NAS methods.}
	\label{algo:framework}
\end{algorithm}

\vspace{1em}

\textbf{Comparison with Prior Works}
\begin{itemize}[leftmargin=*]
    \setlength\itemsep{0.1em}
    \item Mellor et al. \cite{mellor2021neural} only leveraged sample-wise correlation of Jacobian, with no detailed explanation of which aspect (trainability/expressivity/generalization) this indicator represents. Moreover, they only leveraged Random Search on NAS-Bench-201, without studying more NAS methods and search spaces.
    \item Abdelfattah et al. \cite{abbdelfattah2020zero} mainly leveraged ``synflow'' indicator equipped with ``warm-up'' or ``move proposal'' search strategy, which is related to trainability. However, they still have to use trained models for proxy inference during the search.
    \item Chen et al. \cite{chen2020tenas} built their framework on top of a super-net based approach, and strongly rely on a highly customized super-net pruning strategy. We evaluate their method without pruning (shown in Table \ref{table:nasbench201}) and observed inferior performance.
\end{itemize}

\newcolumntype{C}{ >{\centering\arraybackslash} m{4.9cm} }
\newcolumntype{D}{ >{\centering\arraybackslash} m{0.2cm} }
\begin{figure*}[!t]
    \centering
    \renewcommand{\arraystretch}{0.1}
    \begin{tabular}{DDCCC}
        \toprule
        \multirow{2}{*}[-2em]{\rotatebox[origin=c]{90}{NAS-Bench-201}} & \rotatebox[origin=l]{90}{\underline{Early} Search Stage} &
        \includegraphics[width=0.3\textwidth]{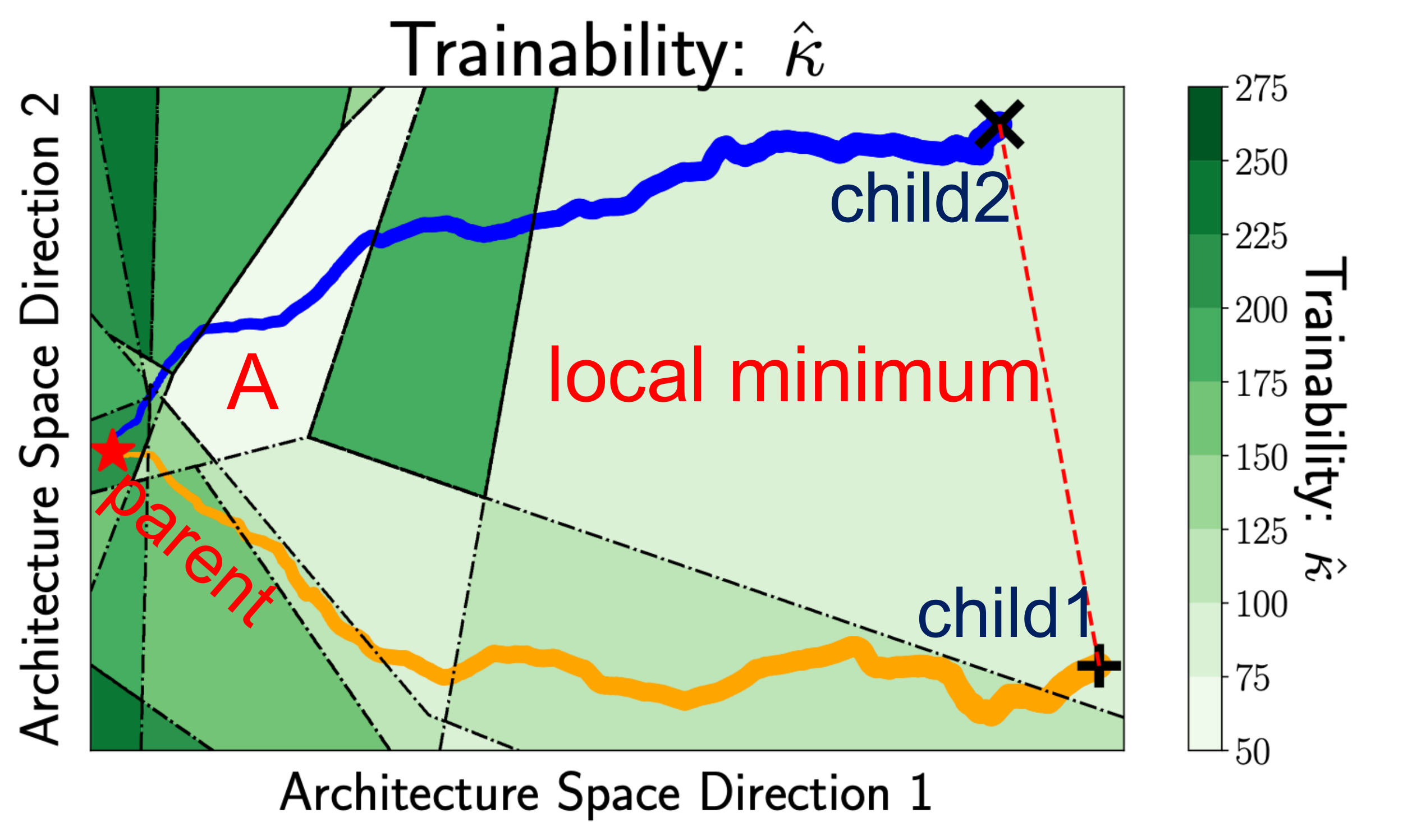} &
        \includegraphics[width=0.3\textwidth]{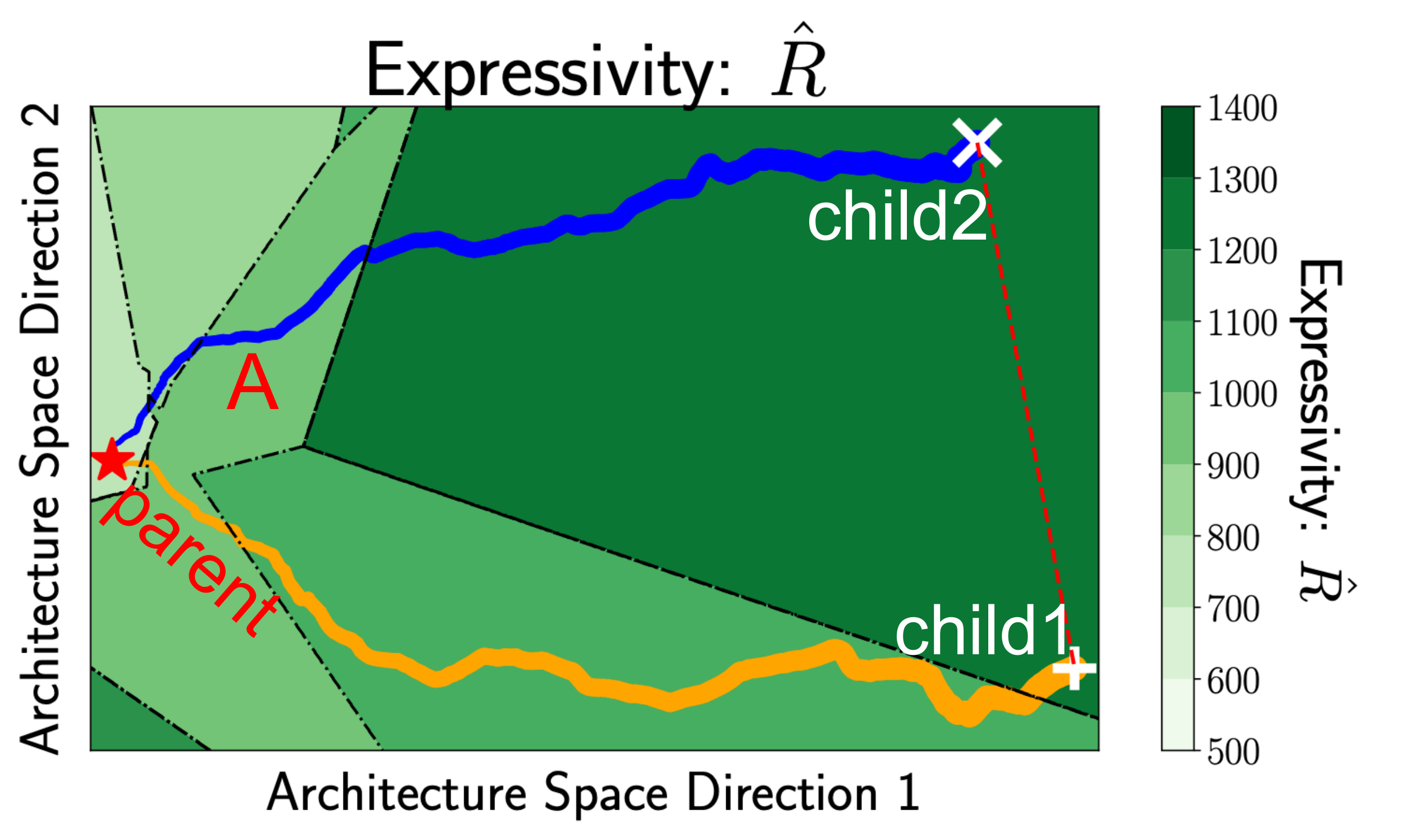} &
        \includegraphics[width=0.3\textwidth]{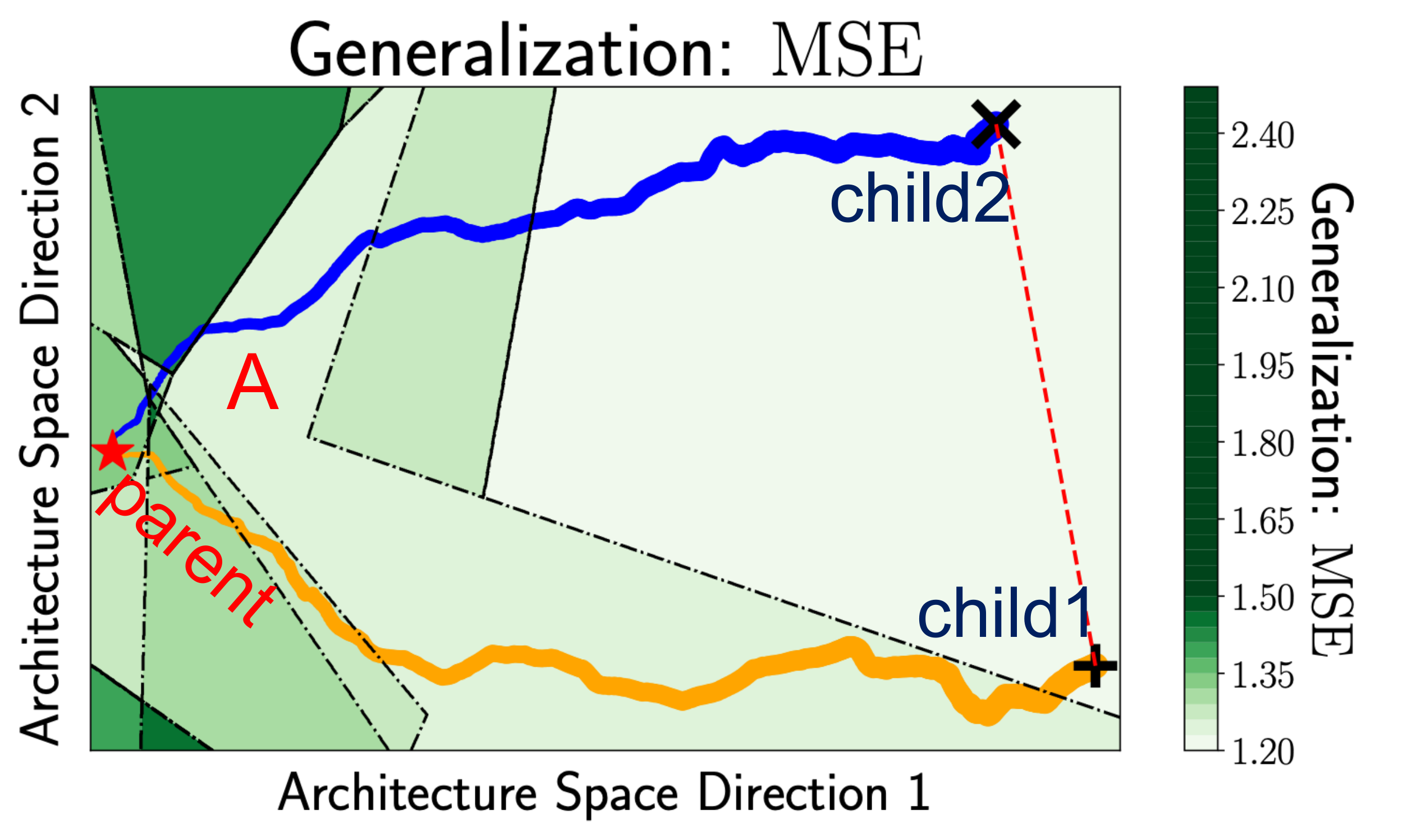}
        \\ \cmidrule{2-5}
        
         & \rotatebox[origin=l]{90}{\underline{Late} Search Stage} &
        \includegraphics[width=0.3\textwidth]{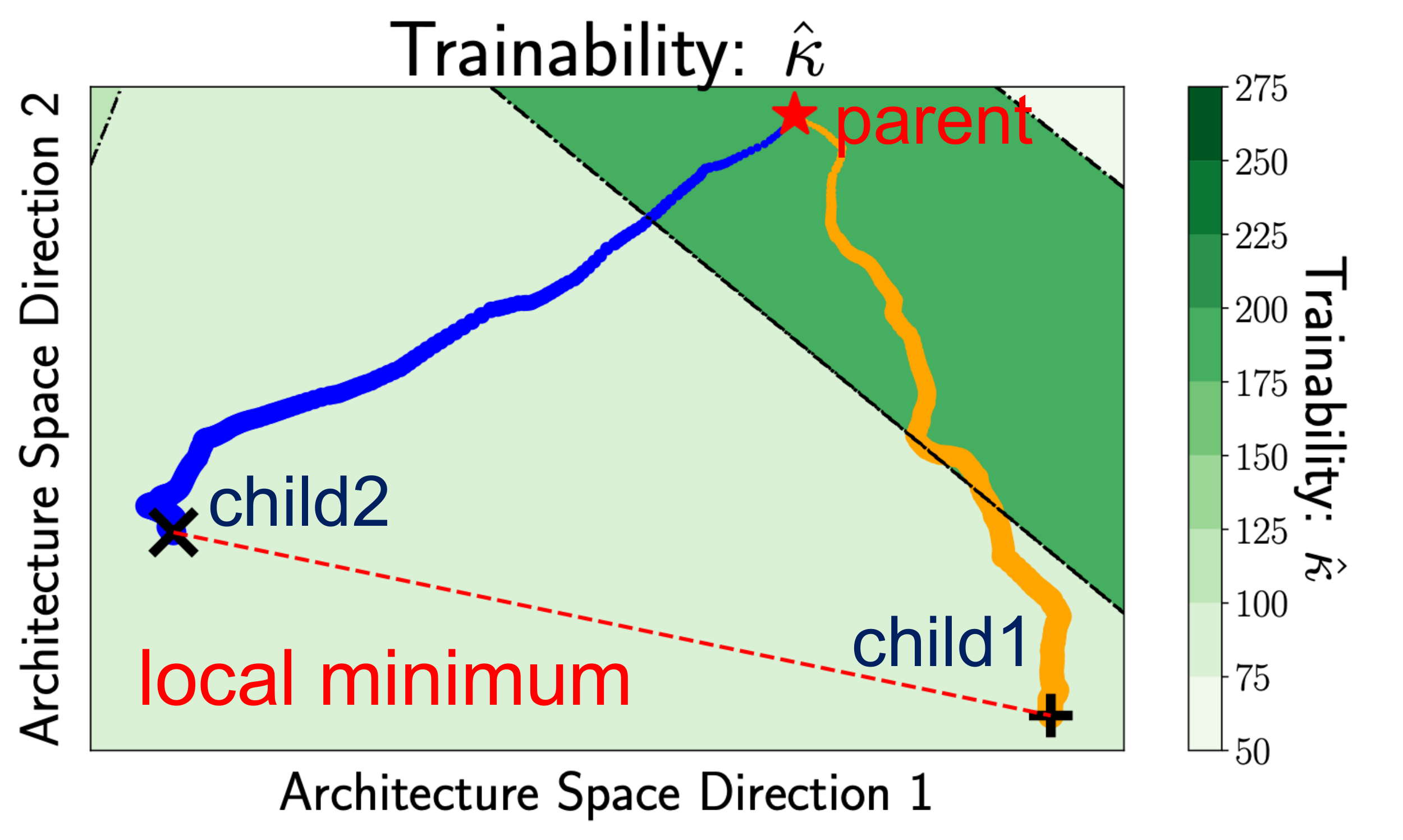} &
        \includegraphics[width=0.3\textwidth]{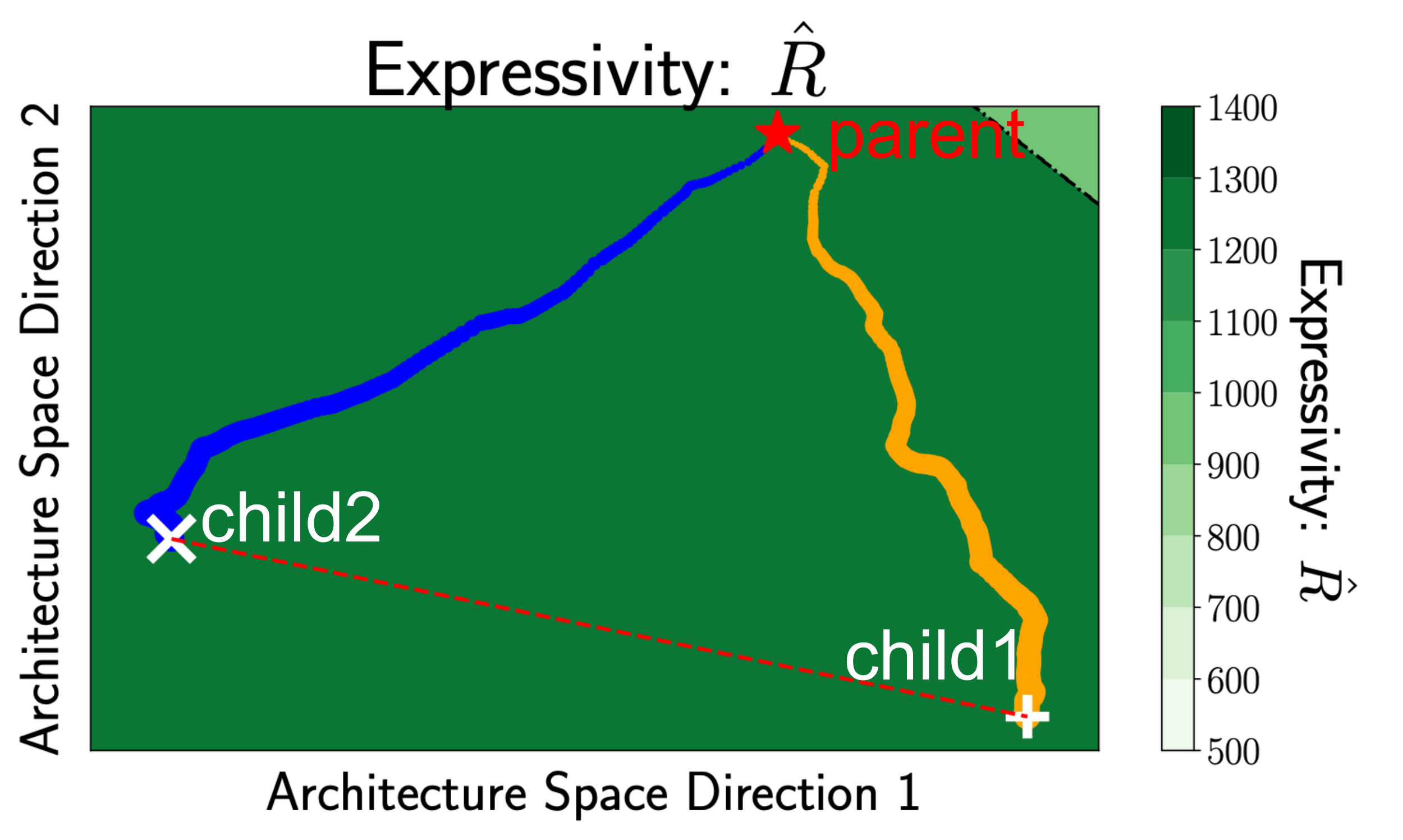} &
        \includegraphics[width=0.3\textwidth]{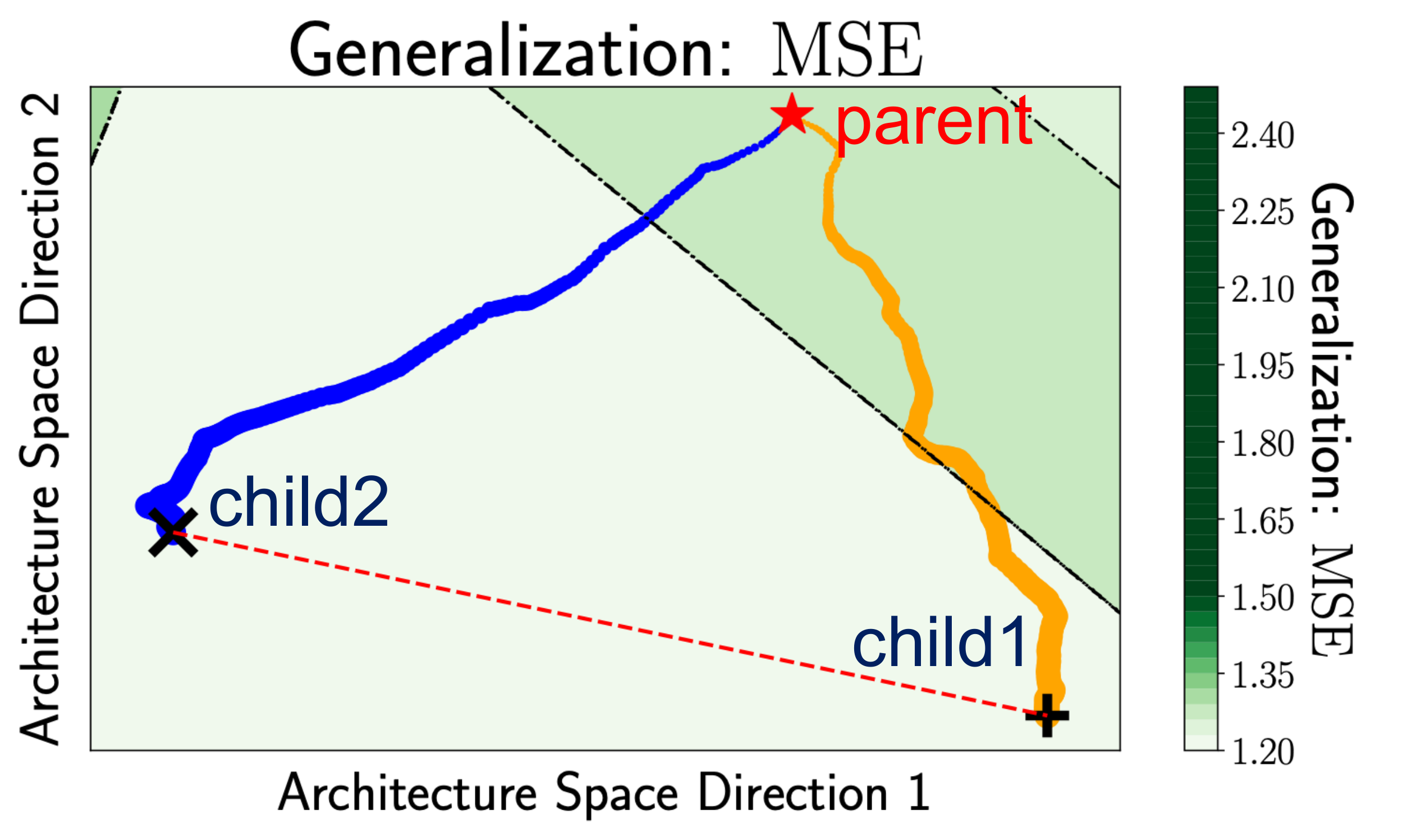}
         \\ \midrule
         
        \multirow{2}{*}[-2em]{\rotatebox[origin=c]{90}{DARTS Space}} & \rotatebox[origin=l]{90}{\underline{Early} Search Stage} & 
        \includegraphics[width=0.3\textwidth]{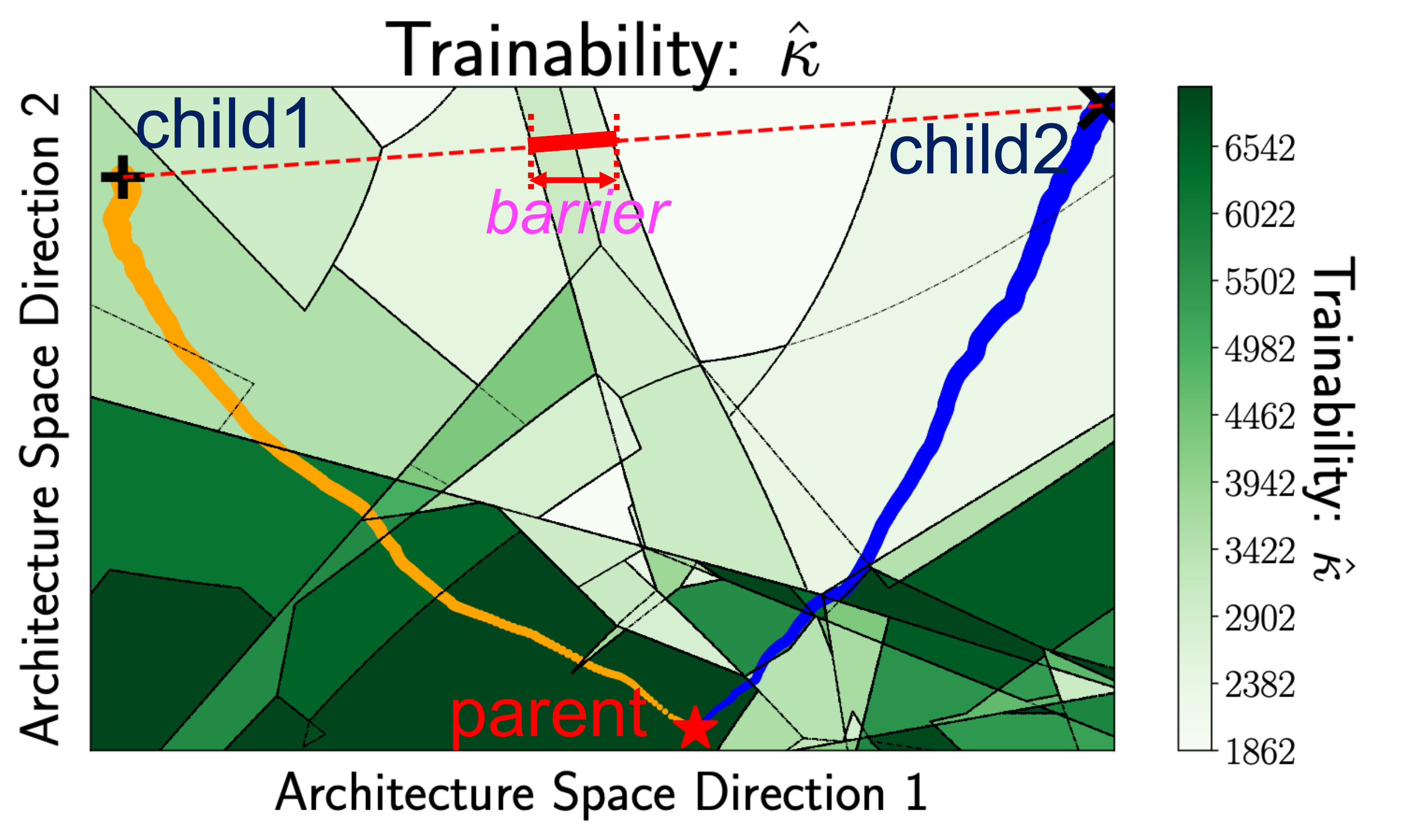} &
        \includegraphics[width=0.3\textwidth]{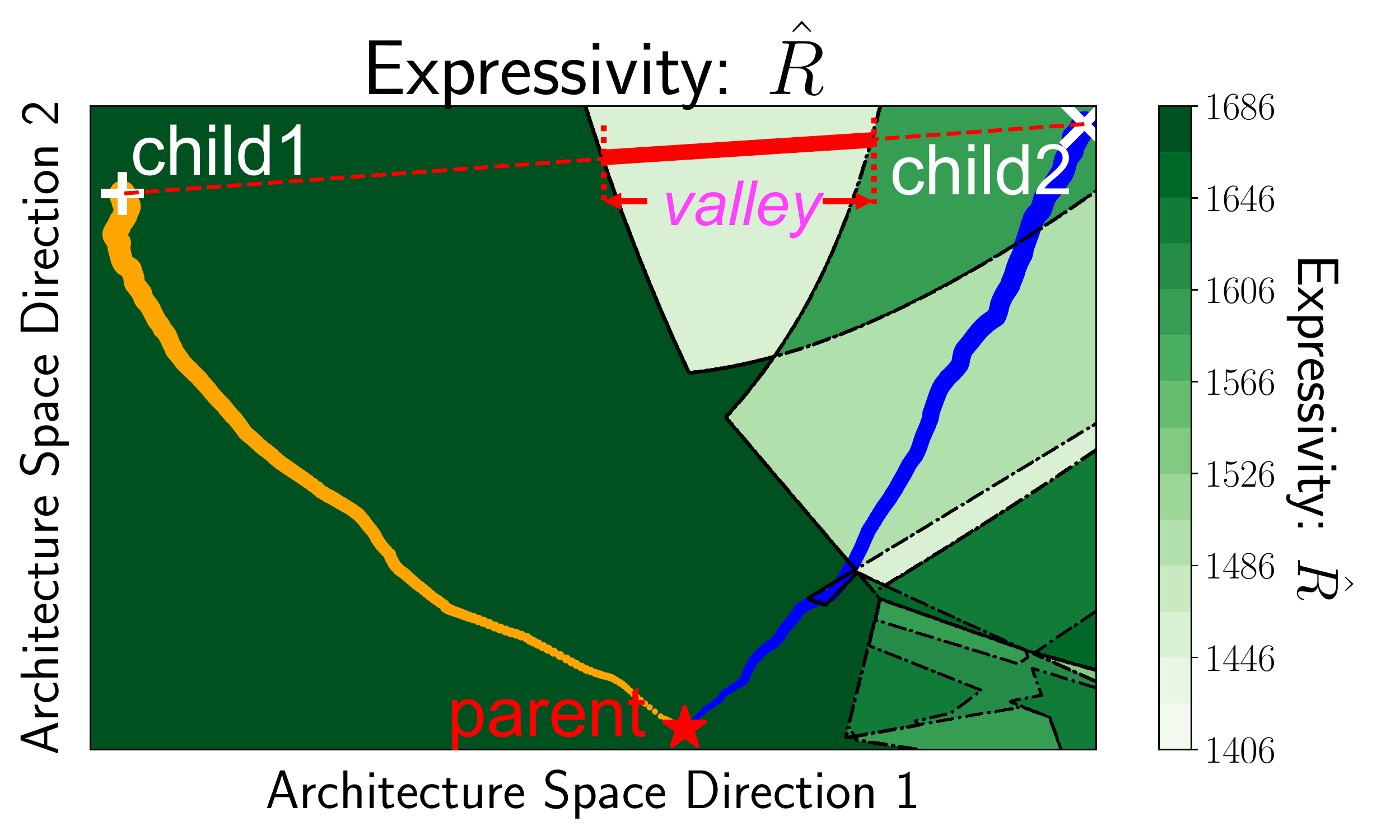} &
        \includegraphics[width=0.3\textwidth]{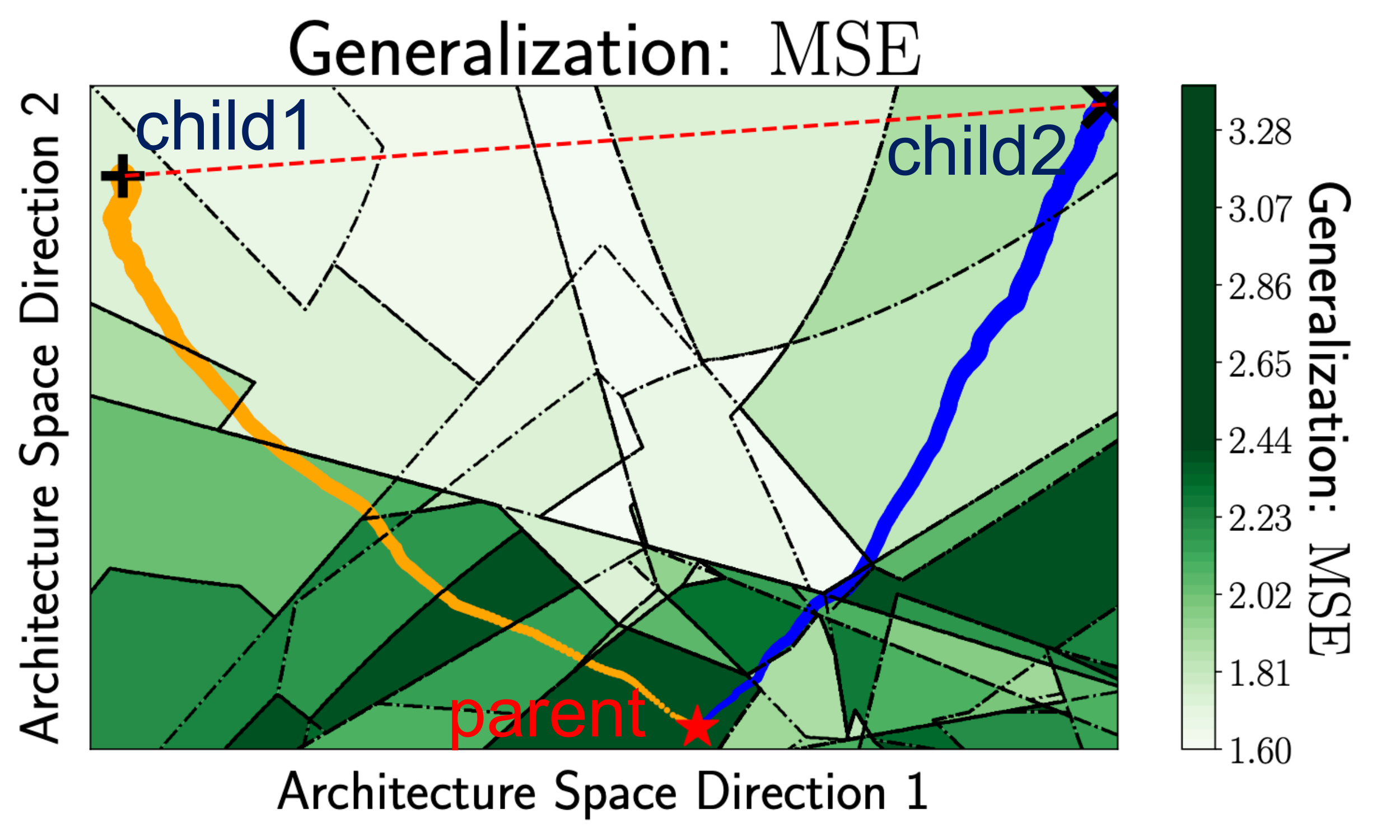}
        \\ \cmidrule{2-5}
        
         & \rotatebox[origin=l]{90}{\underline{Late} Search Stage} &
        \includegraphics[width=0.3\textwidth]{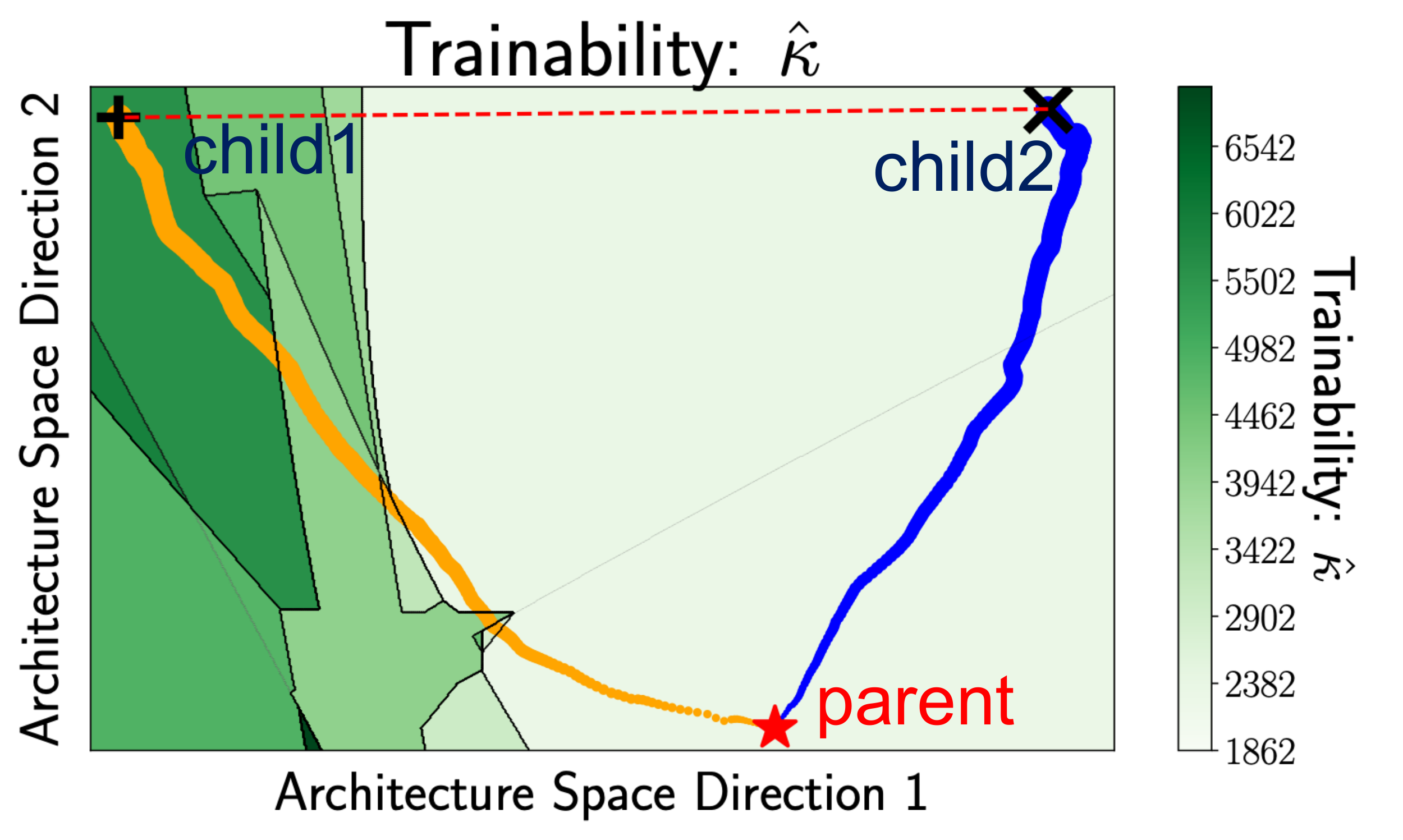} &
        \includegraphics[width=0.3\textwidth]{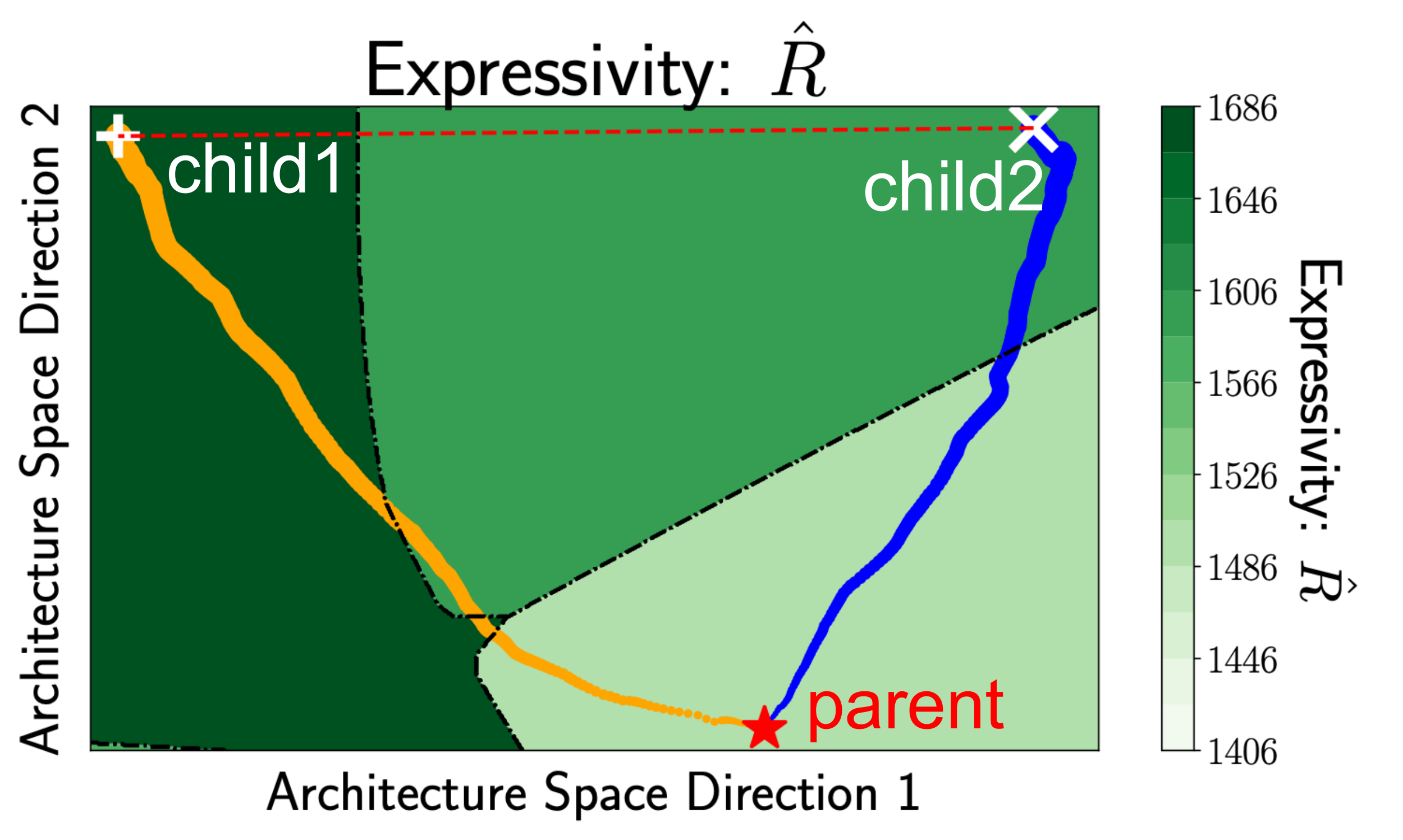} &
        \includegraphics[width=0.3\textwidth]{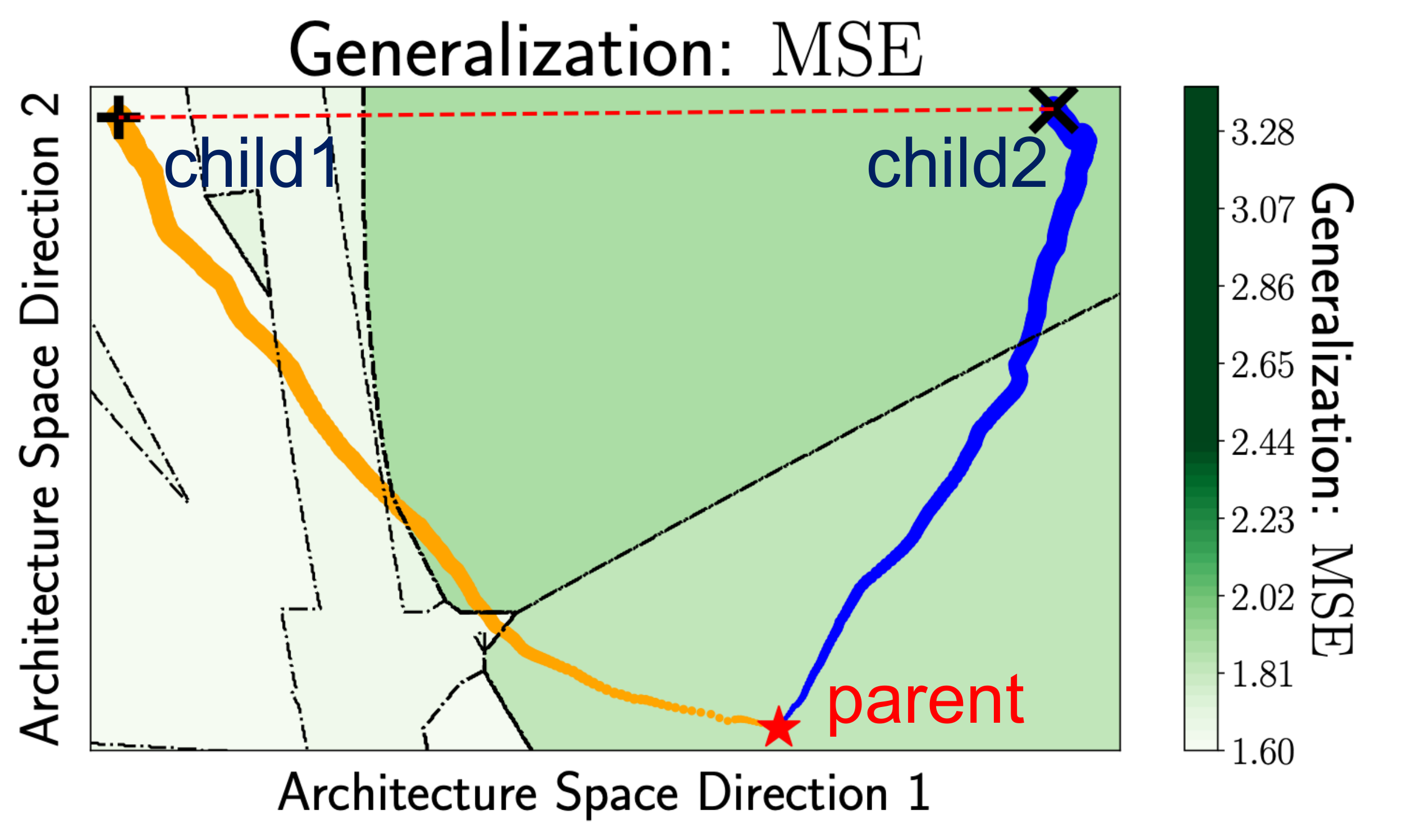}
         \\ \bottomrule
    \end{tabular}
    \caption{The architecture landscape with respect to trainability (left), expressivity (middle), and generalization (right) on a 2D plane projected via PCA from the search space. The red star indicates the parent of a search by Reinforcement Learning, and two crosses represent two children spawned from the same parent (at ``Early'' or ``Late'' search stage), but searched with different randomness. The simple geometry of NAS-Bench-201 is easier to search, whereas the DARTS space is much more complex and hard to explore.}\label{fig:landscape}
\end{figure*}

\subsection{Visualizing Search Process on Different Architecture Landscapes} \label{sec:landscapes}
It has been a missing part in the NAS community to visualize the search process on architecture landscapes from different search spaces. Several bottlenecks hinder this analysis: 1) evaluation via truncated training still suffers from heavy computation cost, making the architecture landscape intractable to characterize; 2) the truncated accuracy or loss value is noisy, making the trade-off of exploration-exploitation of the search process hard to observe.

We leverage the Reinforcement Learning (RL) as the example, and take pioneering steps to conduct such analysis:
\begin{itemize}[leftmargin=*]
    \setlength\itemsep{0.1em}
    \item To explore the global and local geometry of the architecture landscapes, at search step $t$ we spawn a parent architecture into two children, and proceed the search of these two children with different randomness.
    \item We collect the trajectory of architectures from two children, and project the high-dimensional architecture space (represented by the categorical policy distribution of the RL agent) into a 2D plane via PCA.
\end{itemize}
We perform these analysis at early and late search steps on both NAS-Bench-201 space \cite{dong2020bench} and DARTS search space \cite{liu2018darts} (see search space details in Section \ref{sec:exp}), shown in Figure \ref{fig:landscape}. Our observations are summarized as follows:

\begin{itemize}[leftmargin=*]
    \setlength\itemsep{0.1em}
    \item On NAS-Bench-201, the search can land in areas where $\hat{\kappa}_t$, $\hat{R}_t$, and $\mathrm{MSE}_t$ are all good. Although some areas (e.g. area ``A'' in early NAS-Bench-201) enjoy local minima on one of the three aspects, the search will proceed beyond it due to its inferiority on the other two properties.
    \item Spawning at both early and late search stage, two children from NAS-Bench-201 land in the same area in the end. This is probably because of the simple operator types and topology from the design of NAS-Bench-201 (see details in Section \ref{sec:201} and original paper \cite{dong2020bench}).
    \item DARTS space is associated with much more complex architecture landscapes. Children spawned from both early and late stage may land in different areas, with a barrier (or a valley) on their interpolation (orange dashed line). This complex landscape introduce significant challenge to {balance and trade-off} $\hat{\kappa}_t$, $\hat{R}_t$, and $\mathrm{MSE}_t$ for NAS search, with even noisy signals.
\end{itemize}
In general, our landscape analysis on the architecture space can be analogized to the counterpart on the parameter space~\cite{li2017visualizing}. The architectural landscape can influence the behavior of the architecture search.
Our visualizations help explore the sharpness/flatness of architectural minimizers found by different search methods, in different architecture spaces, and the choices of different architectural compositions (skip connections, channel numbers, network depths, etc.).
Specifically, the usage and impact of our landscape analysis are explained below:
\begin{enumerate}[leftmargin=*]
    \setlength\itemsep{0.1em}
    \item Compare different search spaces: given the same search method, the search trajectory on different search spaces will lead to different behaviors. If a search space incurs plenty of barriers or valleys on the trajectory, that means it poses challenges for the search method to converge.
    \item Compare different search algorithms: given the same search space, besides the final search performance, a good search algorithm should be able to explore a vast amount of areas in a search space, instead of being trapped in local regions.
    \item Design search space: \cite{radosavovic2019network,radosavovic2020designing}, visualizing the architecture landscapes can help study the complexity and geometry of the search space and avoid rough architecture landscapes. Specifically, we can evaluate compositions in a search space, by comparing the change in the landscape when we add different operators or connections.
    \item Design search algorithm: visualizing the search process can reveal the quality and stability of the search via different spawning and randomness.
\end{enumerate}

\begin{table*}[!htp]
\centering
\caption{Search Performance on NAS-Bench-101. We ran TEG-NAS for 10 times and report the mean test accuracy and std.}
\begin{threeparttable}
\scalebox{0.92}{
\begin{tabular}{lrrcccrr}
\toprule
 Method & GPU Hours & \#Queries &  Test Acc.(\%) & STD(\%) & Test Regret(\%)  & Avg. Rank & Search
Method \\
\midrule
LaNAS \cite{wang2019sampleefficient} & 107.3\textsuperscript{$\dagger$} & 200 & 93.90 & - & 0.42 & 168.1 & Sample-based\\
BONAS \cite{shi2019bridging} & 107.3\textsuperscript{$\dagger$} & 200 & 94.09 & - & 0.23 & 18.0 & Sample-based\\
NASBOWLr \cite{ru2020interpretable} & 80.5\textsuperscript{$\dagger$} & 150 & 94.09 & - & 0.23 & 18.0 & Sample-based\\
CATE (cate-DNGO-LS) \cite{yan2021cate} & 80.5\textsuperscript{$\dagger$} & 150 & 94.10 & - & 0.22 & 12.3 & Sample-based\\
WeakNAS \cite{wu2021stronger} & 80.5\textsuperscript{$\dagger$} & 150 & 94.10 & 0.19 & 0.22 & 12.3 & Sample-based\\
\midrule
ZERO-COST NAS\cite{abbdelfattah2020zero}\textsuperscript{$\ddagger$} & 27.3\textsuperscript{$\dagger$} & 51 & 94.22 & - & 0.10 & 3.0 & Training-Free + Sample-based\\
\midrule
Synflow \cite{tanaka2020pruning} & - & - & 91.31 & 0.02 & 3.01 & 156663.0 & Training-Free\\
NASWOT \cite{mellor2021neural} & 0.006 & - & 91.77 & 0.05 & 2.55 & 118291.0 & Training-Free\\
AREA (Evolution + NASWOT) \cite{mellor2021neural} & 3.33 & - & 93.91 & 0.29 & 0.41 & 153.0 & Training-Free\\
GenNAS-N \cite{li2021generic} & 5.75 & - & 93.92 & 0.004 & 0.40 & 135.0 & Training-Free\\
Evolution & 2.22 & 170 & 92.17 & 2.19 & 2.15 & 85891.0 & Sample-based\\
\textbf{Evolution + TEG (ours)} & 0.78 & 250 & 92.52 & 1.30 & 1.80 & 59676.0 & Training-Free\\
REINFORCE & 2.77 & 200 & 93.80 & 0.12 & 0.52 & 441.0 & Sample-based \\
\textbf{REINFORCE + TEG (ours)} & 0.24 & 250 & 94.11 & 0.11 & 0.21 & 12.0 & Training-Free \\

\midrule
 Optimal & - & - & 94.32 & - & 0.00 & 1.0 & - \\
\bottomrule
\end{tabular}}

    \begin{tablenotes}
        \item[$\dagger$] Estimated results via the number of queries, where each query in NAS-Bench-101 takes an average of 1932s to train from scratch.
        \item[$\ddagger$] ZERO-COST NAS~\cite{abbdelfattah2020zero} use training-free metrics to warm-up and initialize the sampled-based search algorithm, thus is considered a hybrid of both.
    \end{tablenotes}
\end{threeparttable}

\label{table:nasbench101}
\end{table*}

\begin{table*}[!t]
    \centering
    \caption{Search Performance from NAS-Bench-201. Test accuracy with mean and deviation are reported. ``optimal'' indicates the best test accuracy achievable in the space. The search time cost of our TEG-NAS is agnostic to the size of dataset (Section \ref{sec:framework}). For REINFORCE \cite{williams1992simple}, Evolution \cite{real2019regularized}, and FP-NAS \cite{fpnas}, three search costs are listed for CIFAR-10/CIFAR-100/ImageNet-16-120.}
    \resizebox{0.95\textwidth}{!}{
    \begin{tabular}{lcccccc}
    \toprule
    \textbf{Architecture} & \textbf{CIFAR-10} & \textbf{CIFAR-100} & \textbf{ImageNet-16-120} & \textbf{\tabincell{c}{Search Cost\\(GPU sec.)}} & \textbf{\tabincell{c}{Search\\Method}} \\ \midrule
    ResNet \cite{he2016deep} & 93.97 & 70.86 & 43.63 & - & - \\ \midrule
    RSPS \cite{li2020random} & $87.66(1.69)$ & $58.33(4.34)$ & $31.14(3.88)$ & 8007.13 & random \\
    ENAS \cite{pham2018efficient} & $54.30(0.00)$ & $15.61(0.00)$ & $16.32(0.00)$ & 13314.51 & RL \\
    DARTS (1st) \cite{liu2018darts} & $54.30(0.00)$ & $15.61(0.00)$ & $16.32(0.00)$ & 10889.87 & gradient \\
    DARTS (2nd) \cite{liu2018darts} & $54.30(0.00)$ & $15.61(0.00)$ & $16.32(0.00)$ & 29901.67 & gradient \\
    GDAS \cite{dong2019searching} & $93.61(0.09)$ & $70.70(0.30)$ & $41.84 (0.90)$ & 28925.91 & gradient \\
    DrNAS \cite{chen2020drnas} & $ 94.36 (0.00)$ & $  73.51 (0.00)$ & $ 46.34 (0.00)$ & - & gradient \\
    RLNAS \cite{zhang2021neural} & $93.45$ & $70.71$ & $43.70$ & - & gradient \\
    G-EA \cite{lopes2021guided} & $93.98 (0.18)$ & $ 72.12 (0.35)$ & $45.94 (0.71)$ & 18567 & gradient \\
    $\beta$-DARTS \cite{ye2022beta} & $ 94.36 (0.00)$ & $  73.51 (0.00)$ & $ 46.34 (0.00)$ & 11520 & gradient \\
    Single-DARTS \cite{hou2021single} & $ 94.36 (0.00)$ & $  73.51 (0.00)$ & $ 46.34 (0.00)$ & - & gradient \\
    NASWOT ($N=1000$) \cite{mellor2021neural} & $92.96 (0.81)$ & $69.98 (1.22)$ & $44.44(2.10)$ & 306.19 & training-free \\
    TE-NAS \cite{chen2020tenas} & $93.9(0.47)$ & $71.24(0.56)$ & $42.38(0.46)$ & 1558 & training-free \\
    \midrule
    TE-NAS + TEG (ours) & $93.94(0.2)$ & $71.44(0.81)$ & $44.11(0.88)$ & 3330.5 & training-free \\
    REINFORCE \cite{williams1992simple} & $90.00(1.16)$ & $68.40 (5.93)$ & $44.78(1.20)$ & 33.9k/35.9k/63.5k & RL \\
    REINFORCE + TEG (ours) & $90.21(0.67)$ & $70.42 (0.36)$ & $44.88(0.91)$ & 3668.5 & training-free \\
    Evolution \cite{real2019regularized} & $90.92(0.31)$ & $69.32(3.31)$ & $44.33(1.81)$ & 33.7k/38.1k/148.3k & evolution \\
    Evolution + TEG (ours) & $91.00(0.33)$ & $70.10(1.47)$ & $44.45(0.75)$ & 9939.5 & training-free \\ 
    FP-NAS \cite{fpnas} & $ 55.38 (1.52)$ & $ 22.30 (15.45)$ & $ 16.96(6.43)$ & 3.7k/6.6k/17.2k & gradient \\
    FP-NAS + TEG (ours) & $ 93.73(0.50)$ & $70.36(0.44)$ & $ 46.03(0.10)$ & 641.67 & training-free \\\midrule
    \textbf{Optimal} & 94.37 & 73.51 & 47.31 & - & - \\
    \bottomrule
    \end{tabular}\label{table:nasbench201}}
\end{table*}

\section{Experiments} \label{sec:exp}

Following the experimental setting in~\cite{abbdelfattah2020zero,mellor2021neural,zhang2021neural,lopes2021guided,ye2022beta,hou2021single,chen2020drnas,chen2020tenas},
in this section, we evaluate our TEG-NAS framework on three commonly used search spaces: NAS-Bench-101~\cite{pmlr-v97-ying19a}, NAS-Bench-201~\cite{dong2020bench}, and DARTS~\cite{liu2018darts}.
For DARTS space, we conduct experiments on both CIFAR-10 and ImageNet (Section~\ref{sec:cifar10_imagenet}). For NAS-Bench-201, we test all three supported datasets (CIFAR-10, CIFAR-100, ImageNet-16-120 \cite{imagenet16}) in section \ref{sec:201}.

\subsection{Studied Search Methods} \label{sec:nas_method_comparison}

\textbf{REINFORCE} \cite{williams1992simple,pham2018efficient}: Reinforcement learning (RL) treats the NAS search process as a sequential decision making process. The policy agent formulates a single-path architecture by choosing a sequence of operators as actions, and uses the accuracy, loss value, or our training-free indicators of sampled architecture as the reward to update its internal policy distribution.

\textbf{Evolution} \cite{real2019regularized}: Starting from a randomly initialized pool of architectures, the evolution keeps updating the population by mutating the high-ranked architectures. The ranking criteria could be the accuracy, loss value, or our training-free indicators of sampled architectures.

\textbf{Fast Probabilistic NAS}: FP-NAS views search evaluations from an underlying distribution over architectures \cite{fpnas}. It constructs a supernet and corresponding architecture parameters. Leveraging Importance Weighted Monte-Carlo EB algorithm\cite{carlin2000empirical}, architecture parameters are optimized to maximize the model likelihoods of sampled architectures, which are weighted by a proxy architecture performance indicator, like accuracy, loss value, or training-free indicators.

\subsection{Implementation Details}

In this section, we include more details regarding Algorithm~\ref{algo:framework} in terms of different search methods.

\subsubsection{Reinforcement Learning}\label{sec:supp_rl}

The policy agent maintains a internal state to represent the architecture search space, denoted as $\bm{\theta}^{\bm{A}}$. This internal state can be converted to a categorical distribution of the architectures ($\bm{\mathcal{A}}$) via softmax: $\bm{\mathcal{A}} = \sigma(\bm{\theta}^{\bm{A}})$.

\textbf{Stopping Criterion}: We stop the RL search when the entropy of $\bm{\mathcal{A}}$ stops decreasing (total iterations $T = 500$ in our work). We train the RL agent with a learning rate as $\eta = 0.04$ on NAS-Bench-201 and $\eta = 0.07$ on DARTS space.

\textbf{Architecture Sampling}: In each iteration, the agent samples one architecture $a_t$ from $\bm{\mathcal{A}}$.

\textbf{Update}: We update the RL agent via policy gradients.
\begin{align}
    \bm{\theta}^{\bm{A}}_{t+1} = \bm{\theta}^{\bm{A}}_{t} &- \eta \cdot \nabla_{\bm{\theta}^{\bm{A}}} f(\bm{\theta}^{\bm{A}}_{t}) \quad t = 1,\cdots,T \\
    f(\bm{\theta}^{\bm{A}}_{t}) &= -\mathrm{log}(\sigma(\bm{\theta}^{\bm{A}})) \cdot (r_t - b_t)\\
    b_t = \gamma b_{t-1} &+ (1 - \gamma) r_t \quad (b_0 = 0, \gamma = 0.9)
\end{align}
$r$ stands for reward, and $b$ for an exponential moving average of reward for the purpose of variance reduction. For the baseline method, reward is taken from the proxy inference, i.e., the test accuracy by 1-epoch truncated training. For our TEG-NAS, reward is composited of three parts: $r = r^\kappa + r^R + r^\mathrm{MSE}$, and we show the justification for how we combine the our indicators in Table~\ref{table:ablation}. Taking $r^\kappa$ (reward from trainability) as the example:
\begin{align}
    r^{\kappa}_t &= \frac{\hat{\kappa}_t - \hat{\kappa}_{t-1}}{\hat{\kappa}_{\mathrm{max},t} - \hat{\kappa}_{\mathrm{min},t}}\\
    \hat{\kappa}_{\mathrm{max},t} &= \mathrm{max}(\hat{\kappa}_1, \hat{\kappa}_2, \cdots, \hat{\kappa}_t)\\
    \hat{\kappa}_{\mathrm{min},t} &= \mathrm{min}(\hat{\kappa}_1, \hat{\kappa}_2, \cdots, \hat{\kappa}_t)
\end{align}
where $\hat{\kappa}_t$ is the evaluated trainability of the architecture sampled at step $t$. We calculate $r^R$ (expressivity) and $r^\mathrm{MSE}$ (generalization) in the same way.

\textbf{Architecture Deriving}: To derive the final searched network, the agent chooses the architecture that has the highest probability, i.e., $a^* = \mathrm{argmax}_a \sigma(\bm{\theta}^{\bm{A}})(a)$.

\begin{table*}[!tb]
    \centering
    \caption{Search Performance from DARTS space on CIFAR-10. Our results are averaged over three searched architectures under different random seeds, with standard deviations in parentheses.}
    \scriptsize
    \begin{threeparttable}
    \resizebox{0.75\textwidth}{!}{
    \begin{tabular}{lcccc}
    \toprule
    \textbf{Architecture} & \textbf{\tabincell{c}{Test Error\\(\%)}} & \textbf{\tabincell{c}{Params\\(M)}} & \textbf{\tabincell{c}{Search Cost\\(GPU days)}} & \textbf{\tabincell{c}{Search\\Method}} \\ \midrule
    AmoebaNet-A \cite{real2019regularized} & $3.34$ & 3.2 & 3150 & evolution \\
    PNAS \cite{liu2018progressive}\tnote{$\star$} & $3.41$ & 3.2 & 225 & SMBO \\
    ENAS \cite{pham2018efficient} & 2.89 & 4.6 & 0.5 & RL \\
    NASNet-A \cite{zoph2018learning} & 2.65 & 3.3 & 2000 & RL \\ \midrule
    
    DARTS (1st) \cite{liu2018darts} & $3.00$ & 3.3 & 0.4 & gradient \\
    SNAS \cite{xie2018snas} & $2.85$ & 2.8 & 1.5 & gradient \\
    GDAS \cite{dong2019searching} & 2.82 & 2.5 & 0.17 & gradient \\
    BayesNAS \cite{zhou2019bayesnas} & $2.81$ & 3.4 & 0.2 & gradient \\
    ProxylessNAS \cite{cai2018proxylessnas}\tnote{$\dagger$} & 2.08 & 5.7 & 4.0 & gradient \\
    NASP \cite{yao2020efficient} & 2.83 (0.09) & 3.3 & 0.1 & gradient \\
    P-DARTS \cite{chen2019progressive} & 2.50 & 3.4 & 0.3 & gradient \\ 
    PC-DARTS \cite{xu2019pc} & $2.57$ & 3.6 & 0.1 & gradient \\
    R-DARTS (L2) \cite{zela2019understanding} & 2.95 (0.21) & - & 1.6 & gradient \\
    SGAS (Cri 1. avg) \cite{li2020sgas} & 2.66 (0.24) & 3.7 & 0.25 & gradient \\
    SDARTS-ADV \cite{chen2020stabilizing} & $2.61$ & 3.3 & 1.3 & gradient \\
    DrNAS \cite{chen2020drnas} & 2.46 (0.03) & 4.1 & 0.6\tnote{$\ddagger$} & gradient \\
    $\beta$-DARTS \cite{ye2022beta} & 2.53 (0.08) &  3.75 (0.15) & 0.4 & gradient \\
    Single-DARTS \cite{hou2021single} & 2.46 & 3.3 & - & gradient \\ \midrule
    TE-NAS \cite{chen2020tenas} & $2.63$ & 3.8 & 0.05\tnote{$\ddagger$} & training-free \\
    TE-NAS + TEG (ours) & $2.58 (0.01)$ & 4.8 (0.1) & 0.05\tnote{$\ddagger$} & training-free \\
    REINFORCE \cite{williams1992simple} & $ 3.25 (0.43)$ & 2.4 (0.3) & 1.1\tnote{$\ddagger$} & RL \\
    REINFORCE + TEG (ours) & $ 2.87 (0.04)$ & 4.0 (0.3) & 0.15\tnote{$\ddagger$} & training-free \\
    Evolution \cite{real2019regularized} & $ 3.24 (0.17)$ & 2.1 (0.1) & 2.6\tnote{$\ddagger$} & evolution \\
    Evolution + TEG (ours) & $ 3.16 (0.35)$ & 3.3 (0.5) & 0.4\tnote{$\ddagger$} & training-free \\
    FP-NAS \cite{fpnas} & $ 4.61 (0.56)$ & 2.2 (0.1) & 0.3\tnote{$\ddagger$} & gradient \\
    FP-NAS + TEG (ours) & $ 2.74 (0.18)$ & 4.4 (0.2) & 0.13\tnote{$\ddagger$} & training-free \\
    \bottomrule
    \end{tabular}}
    
    \begin{tablenotes}
        \item[$\star$] No cutout augmentation.
        \item[$\dagger$] Different space: PyramidNet \cite{han2017deep} as the backbone.
        \item[$\ddagger$] Recorded on a single GTX 1080Ti GPU.
    \end{tablenotes}
    \end{threeparttable}
    \label{tab:cifar10}
\end{table*}

\subsubsection{Evolution}

The evolution search is first initialized with a population of 256 architectures by random sampling. We choose this size of the population based on Figure A-1(a) from \cite{real2019regularized}.

\textbf{Stopping Criterion:} We stop the Evolution search when the population diversity stops decreasing (1000 iterations in our work). Population diversity is calculated as the averaged pair-wise architecture difference in their operator types.

\textbf{Architecture Sampling:} Following Real et al. \cite{real2019regularized}, in each iteration, the evolution search first randomly samples a subset of 64 architectures out of the population. We choose this sampling size based on Figure A-1(a) from \cite{real2019regularized}. Next, the best architecture ($a_t$) from this subset is selected. For the baseline method, the best architecture is the top1 ranked by the proxy inference, i.e., the test accuracy by 1-epoch truncated training. For our TEG-NAS, the best architecture is the top1 by the sum of three rankings by trainability, expressivity, and generalization: $\mathrm{rank}^\kappa + \mathrm{rank}^R + \mathrm{rank}^\mathrm{MSE}$.

\textbf{Update:} Following Real et al. \cite{real2019regularized}, in each iteration the population is updated by adding a new architecture and popping out the oldest architecture (the one that stays in the population for the longest time). The new architecture is generated by mutating the sampled one mentioned above. We follow the same mutation strategy from Real et al. \cite{real2019regularized}.

\textbf{Architecture Deriving:} To derive the final searched network, the best architecture from the population is selected, where the criterion is the same as we choose $a_t$ (see above ``Architecture Sampling'').

\subsubsection{Fast Probabilistic NAS}

The Fast Probabilistic NAS (FP-NAS) formulates the search space as a supernet and shares its parameters to its sub-networks. The original FP-NAS search performs alternative optimization between network parameters and the architecture parameters (denoted as $\bm{\theta}^{\bm{A}}$).
This architecture parameter can be converted to a categorical distribution of the architectures ($\bm{\mathcal{A}}$) via softmax: $\bm{\mathcal{A}} = \sigma(\bm{\theta}^{\bm{A}})$.

\textbf{Stopping Criterion}: We stop the FP-NAS search when the entropy of $\bm{\mathcal{A}}$ stops decreasing (total epochs $T = 100$ in our work). We update the architecture parameters with a learning rate as $\eta = 0.1$.

\textbf{Architecture Sampling}: In each step, the FP-NAS samples a subset $\bm{A}_t$ of $\lambda H(\mathrm{Prob}(a_{i}|\bm{\mathcal{A}}))$ architectures from $\bm{\mathcal{A}}$. Here $H$ denotes the distribution entropy and $\lambda$ is a pre-defined scaling factor where we set it to 0.25.

\textbf{Update}: We update the architecture parameters by stochastic gradient descent.
\begin{align}
    \bm{\theta}^{\bm{A}}_{t+1} = \bm{\theta}^{\bm{A}}_{t} &- \eta \cdot \nabla_{\bm{\theta}^{\bm{A}}} f(\bm{\theta}^{\bm{A}}_{t}) \quad t = 1,\cdots,T \\
    f(\bm{\theta}^{\bm{A}}_{t}) = - &\sum_{i=1}^{|\bm{A}_t|} \mathrm{log}(\mathrm{Prob}(a_i|\bm{\mathcal{A}})) \cdot \frac{e^{r_i}}{\sum_{j=1}^{|\bm{A}_t|} e^{r_j}}
\end{align}
$r$ stands for reward, calculated in the same way as we did for Reinforcement Learning (Section \ref{sec:supp_rl}).

\textbf{Architecture Deriving}: To derive the final searched network, the FP-NAS chooses the architecture that has the highest probability, i.e., $a^* = \mathrm{argmax}_a \sigma(\bm{\theta}^{\bm{A}})(a)$.

\begin{table*}[!tb]
    \centering
    \caption{Search Performance from DARTS space on ImageNet. Our results are averaged over three searched architectures under different random seeds, with standard deviations in parentheses.}
    \scriptsize
    \resizebox{0.75\textwidth}{!}{
    \begin{threeparttable}
    \begin{tabular}{lccccc}
    \toprule
    
    \multirow{2}*{\textbf{Architecture}} & \multicolumn{2}{c}{\textbf{Test Error(\%)}} & \multirow{2}*{\textbf{\tabincell{c}{Params\\(M)}}} &
    \multirow{2}*{\textbf{\tabincell{c}{Search Cost\\(GPU days)}}} & 
    \multirow{2}*{\textbf{\tabincell{c}{Search\\Method}}} \\ \cline{2-3}
    
    & top-1 & top-5 & & & \\ \midrule
    
    NASNet-A \cite{zoph2018learning} & 26.0 & 8.4 & 5.3 & 2000 & RL \\
    AmoebaNet-C \cite{real2019regularized} & 24.3 & 7.6 & 6.4 & 3150 & evolution \\
    PNAS \cite{liu2018progressive} & 25.8 & 8.1 & 5.1 & 225 & SMBO \\
    MnasNet-92 \cite{tan2019mnasnet} & 25.2 & 8.0 & 4.4 & - & RL \\ \midrule
    
    DARTS (2nd) \cite{liu2018darts} & 26.7 & 8.7 & 4.7 & 4.0 & gradient \\
    SNAS (mild) \cite{xie2018snas} & 27.3 & 9.2 & 4.3 & 1.5 & gradient \\
    GDAS \cite{dong2019searching} & 26.0 & 8.5 & 5.3 & 0.21 & gradient \\
    BayesNAS \cite{zhou2019bayesnas} & 26.5 & 8.9 & 3.9 & 0.2 & gradient \\
    P-DARTS (CIFAR-10) \cite{chen2019progressive} & 24.4 & 7.4 & 4.9 & 0.3 & gradient \\ 
    P-DARTS (CIFAR-100) \cite{chen2019progressive} & 24.7 & 7.5 & 5.1 & 0.3 & gradient \\
    PC-DARTS (CIFAR-10) \cite{xu2019pc} & 25.1 & 7.8 & 5.3 & 0.1 & gradient \\
    ProxylessNAS (GPU) \cite{cai2018proxylessnas}\tnote{$\dagger$} & 24.9 & 7.5 & 7.1 & 8.3 & gradient \\
    PC-DARTS (ImageNet) \cite{xu2019pc}\tnote{$\dagger$} & 24.2 & 7.3 & 5.3 & 3.8 & gradient \\
    SGAS (Cri 1. avg) \cite{li2020sgas} & 24.42 (0.16) & 7.29 (0.09) & 5.3 & 0.25 & gradient \\
    DrNAS \cite{chen2020drnas}\tnote{$\dagger$} & 23.7 & 7.1 & 5.7 & 4.6 & gradient \\
    RLNAS \cite{zhang2021neural} & 24.0 & 7.1 & 5.7 & - & gradient \\
    $\beta$-DARTS \cite{ye2022beta} & 23.9 & 7.0 & 5.5 & 0.4 & gradient \\
    Single-DARTS \cite{hou2021single} & 23.0 & - & 6.6 & - & gradient \\ \midrule
    TE-NAS \cite{chen2020tenas} & 26.2 & 8.3 & 6.3 & 0.05 & training-free \\
    TE-NAS + TEG (ours) & 23.6 (0.1) & 7.1 (0.03) & 6.6 (0.1) & 0.05 & training-free \\
    REINFORCE \cite{williams1992simple} & 28.2 (1.8) & 9.6 (1.1) & 3.8 (0.4) & 1.1 & RL \\
    REINFORCE + TEG (ours) & 25.1 (0.2) & 7.7 (0.1) & 5.6 (0.3) & 0.15 & training-free \\
    Evolution \cite{real2019regularized} & 29.1 (0.8) & 10.2 (0.5) & 3.3 (0.1) & 2.6 & evolution \\
    Evolution + TEG (ours) & 26.3 (0.6) & 8.4 (0.3) & 4.8 (0.6) & 0.4 & training-free \\
    FP-NAS \cite{fpnas} & 31.2 (1.0) & 11.4 (0.7) & 3.4 (0.1) & 0.3 & gradient \\
    FP-NAS + TEG (ours) & 23.7 (0.2) & 7.0 (0.1) & 6.1 (0.2) & 0.13 & training-free \\
    \bottomrule
    \end{tabular}
    \begin{tablenotes}
        \item[$\dagger$] Architecture searched on ImageNet, otherwise searched on CIFAR-10 or CIFAR-100.
    \end{tablenotes}
    \end{threeparttable}}
    \label{tab:imagenet}
\end{table*}

\subsection{Results on NAS-Bench-101}\label{sec:101}

NAS-Bench-101\cite{pmlr-v97-ying19a} contains 423,624 unique neural architectures exhaustively generated and evaluated from a fixed graph-based search space. The search space is extremely diverse yet expressive, due to its general encoding scheme, consisting of an adjacency matrix and its corresponding operations at each vertex. Specifically, the adjacency matrix is represented by a $7\times 7$ upper-triangular binary matrix, while the operation at each vertex could be any of three operator types: \textit{conv}$1\times 1$, \textit{conv}$3\times 3$ \textit{convolution}, and \textit{average pooling} $3\times 3$.
Each network is trained for 108 epochs and the network’s accuracy at intermediate epoch(s) is also provided. For the baseline methods, the RL agent and Evolution use the validation accuracy after 2-epoch training as the reward or ranking criteria. For all results on NAS-Bench-101, we run for 10 independent times with different random seeds and the mean and standard deviation of test accuracy are reported. Due to the slight difference in test accuracies of architectures, we also include test regret (the absolute accuracy gap to global optimal) and average rank (the ranking distance to global optimal) for a more clear comparison across different search methods.

As shown in Table~\ref{table:nasbench101}, combing our TEG-NAS with REINFORCE or Evolution, we achieve better performance over the baseline.
We significantly boost the searched test accuracy (over 0.3\%$+$) while reducing more than 56\% search time cost.
Note that we did not evaluate supernet-based NAS methods (FP-NAS, TE-NAS, gradient-based NAS) on NAS-Bench-101, since the general graph-based encoding scheme in NAS-Bench-101 makes it incompatible with weight-sharing supernet, which is required in gradient-based NAS.

\subsection{Results on NAS-Bench-201}\label{sec:201}

NAS-Bench-201 \cite{dong2020bench} provides a cell-based search space and the performance of all 15,625 networks it contains using a unified protocol. The network's accuracy is directly available by querying the database, benefiting towards the study of NAS methods without network evaluation.
It contains five operator types: \textit{none} (\textit{zero}), \textit{skip connection}, \textit{conv}$1\times 1$, \textit{conv}$3\times 3$ \textit{convolution}, and \textit{average pooling} $3\times 3$.
We refer to their paper for details of the space.
For the baseline methods, the RL agent and Evolution use the test accuracy after 1-epoch training as the reward or ranking criteria. The FP-NAS uses alternative training between architecture parameters and supernet parameters with stochastic gradient descent.
For all results we report, we run for four independent times with different random seeds, and report the mean and standard deviation in Table \ref{table:nasbench201}.

We can see that for all three NAS methods (REINFORCE, Evolution, FP-NAS), our TEG-NAS framework boosts the search performance while significantly reduce the search time cost.
Moreover, by adopting our unified framework, the accuracy of TE-NAS can be further improved.

\subsection{Results on DARTS Search Space} \label{sec:cifar10_imagenet}

\textbf{Architecture Space}
The DARTS space contains eight operator types: \textit{none} (\textit{zero}), \textit{skip connection}, \textit{separable convolution} $3\times 3 $ and $5\times 5$, \textit{dilated separable convolution} $3\times 3 $ and $5\times 5$, \textit{max pooling} $3\times 3 $, \textit{average pooling} $3\times 3 $.
We stack 20 cells to compose the network and set the initial channel number as 36 \cite{liu2018darts,chen2019progressive,xu2019pc}. We place the reduction cells at the 1/3 and 2/3 of the network. Each cell contains six nodes.

The architecture for ImageNet is slightly different: the network is stacked with 14 cells with the initial channel number set to 48~\cite{xu2019pc,chen2019progressive}.
The spatial resolution is downscaled from $224\times 224$ to $28\times 28$ with the first three convolution layers of stride 2.

\textbf{Evaluation Protocols}\label{sec:implementations}
We follow previous NAS works \cite{xu2019pc,chen2019progressive,chen2020stabilizing} to evaluate architectures after search. On CIFAR-10, we train the searched network with cutout regularization of length 16, drop-path \cite{zoph2018learning} with probability as 0.3, and an auxiliary tower of weight 0.4.
On ImageNet, we also use label smoothing during training.
On both CIFAR-10 and ImageNet, the network is optimized by an SGD optimizer with cosine annealing, with a learning rate initialized as 0.025 and 0.5, respectively.

\textbf{Results}
For example, on ImageNet, our TEG brings improvements for: REINFORCE +3.1\% top-1, -86.4\% time cost; Evolution +2.8\% top-1, -84.6\% time cost; FP-NAS +7.5\% top-1, -56.7\% time cost.
FP-NAS equipped with our TEG framework can even achieve the best top-1 error evaluated on ImageNet. All training-free versions of three NAS methods can now complete the search with less than a half GPU day. These search improvements on the large-scale DARTS space and datasets validate the effectiveness and efficiency of our unified TEG-NAS framework.

We also notice that FP-NAS benefits the most by equipping our TEG method ($+$1.87\% on CIFAR-10 and $+$7.5\% on ImageNet). The underlying problem of ProbNAS is similar to DARTS. As a weight-sharing NAS method, skip-connection favors the gradient flow during search, which introduces a strong bias in the supernet parameters. At the end of search the supernet’s accuracy can not faithfully represent the ranking of single-path networks. This problem is pointed out in recent NAS works~\cite{chen2020stabilizing,yu2020train}. In contrast, our training-free method can address this problem: we avoid any gradient descent, and the shared weight (at its initialization) will not be affected by any inductive bias during training, thus unleashing more power of weight-sharing NAS methods.

\begin{table}[!b]
    \centering
    \caption{Ablation study of different combinations of training-free indicators for REINFORCE NAS method on NAS-Bench-201 CIFAR-100. Test accuracy with mean and deviation are reported.}
    \resizebox{0.46\textwidth}{!}{
    \begin{tabular}{ccc}
        \toprule
        Indicators & Accuracy & GPU secs. \\ \midrule
        Baseline (1-epoch training) & 68.4(5.93) & 19786 \\
        $\hat{R}$ & 69.06(2.15) & 254.6 \\
        $\hat{\kappa}$ & 69.27(0.73) & 1574 \\
        $\mathrm{MSE}$ & 69.68(0.88) & 2447.7 \\
        $\hat{\kappa}$, $\hat{R}$ & 69.89(0.98) & 1716.1 \\
        $\hat{R}$, $\mathrm{MSE}$ & 70.00(0.42) & 2667.4 \\
        $\hat{\kappa}$, $\mathrm{MSE}$ & 70.18(0.04) & 2704 \\
        $\hat{\kappa}$, $\hat{R}$, $\mathrm{MSE}$ & 70.42(0.36) & 3668.5 \\
        \bottomrule
    \end{tabular}\label{table:ablation}}
\end{table}

\subsection{Searched Architecture on DARTS Search Space}

\label{app:vis}
\begin{figure}[!htb]
\centering
\subfigure[Normal Cell]{\includegraphics[width=0.49\linewidth]{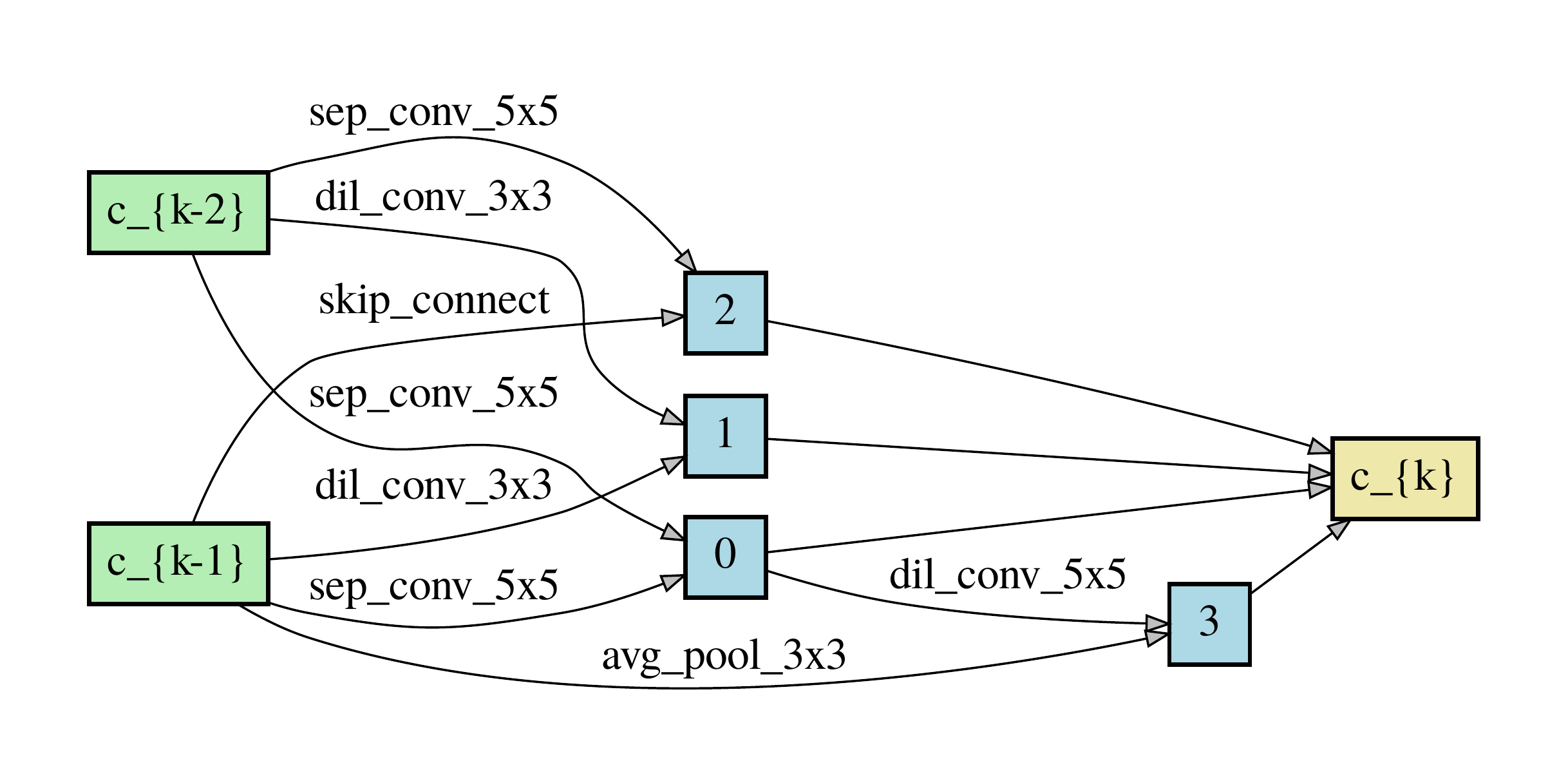}}
\hfill
\subfigure[Reduction Cell]{\includegraphics[width=0.49\linewidth]{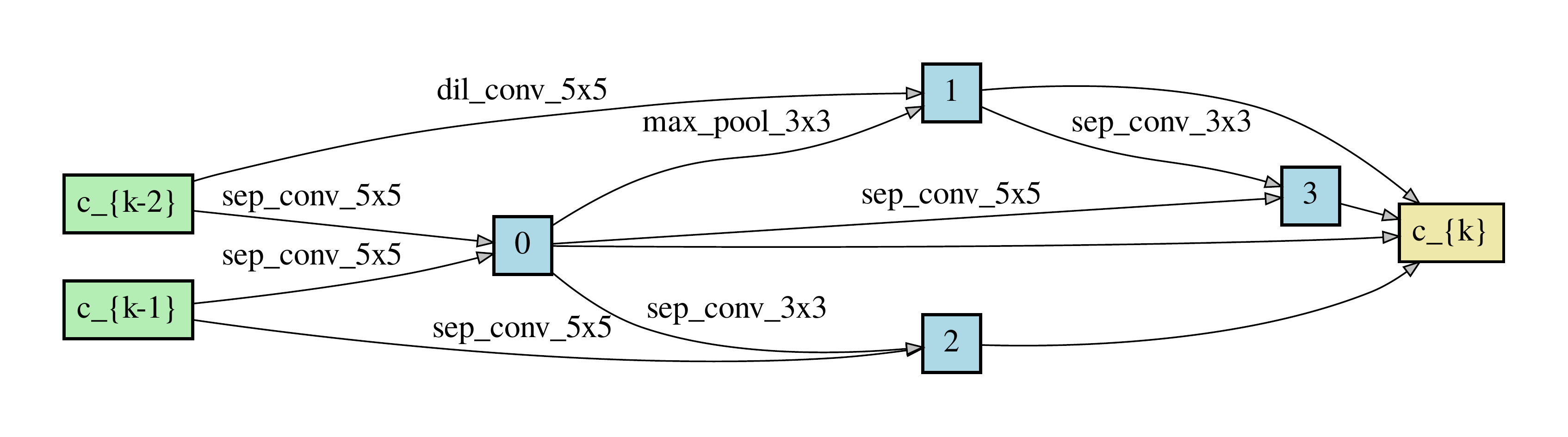}}
\caption{Normal and Reduction cells discovered by RL + TEG-NAS on DARTS space on CIFAR-10.}
\label{fig:RL_TEG_cifar10}
\end{figure}

\begin{figure}[!htb]
\centering
\subfigure[Normal Cell]{\includegraphics[width=0.49\linewidth]{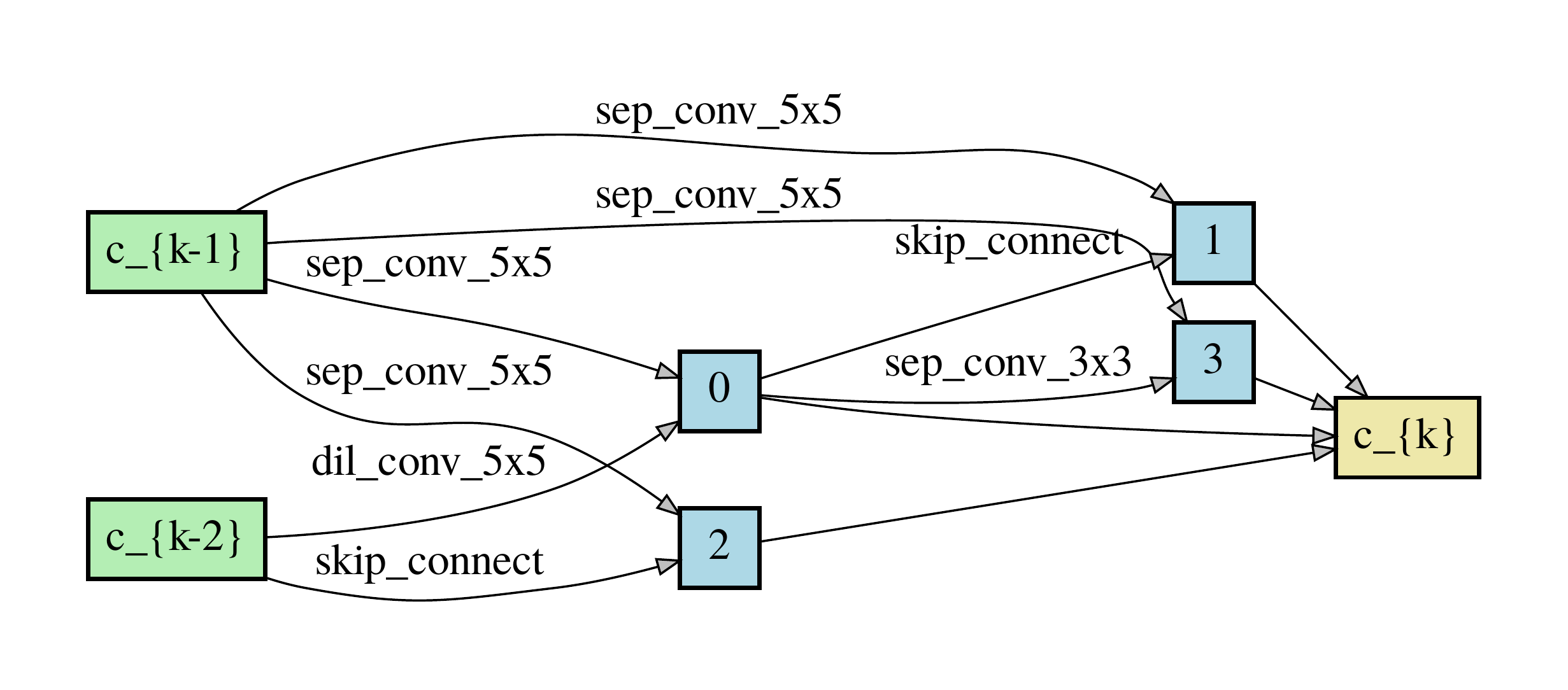}}
\hfill
\subfigure[Reduction Cell]{\includegraphics[width=0.49\linewidth]{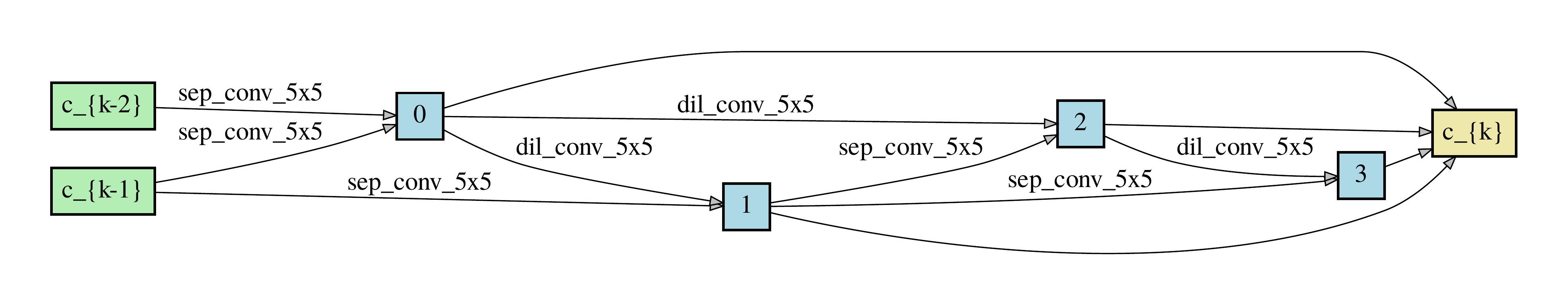}}
\caption{Normal and Reduction cells discovered by Evolution + TEG-NAS on DARTS space on CIFAR-10.}
\label{fig:Evolution_TEG_cifar10}
\end{figure}

\begin{figure}[!htb]
\centering
\subfigure[Normal Cell]{\includegraphics[width=0.49\linewidth]{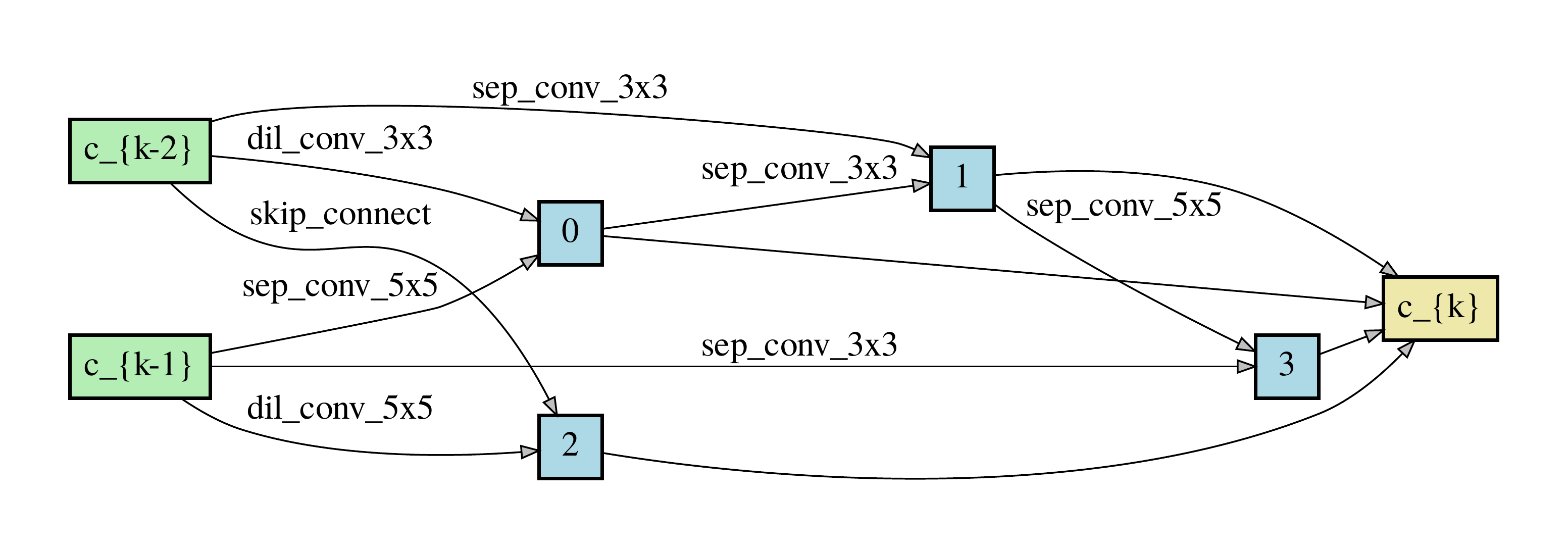}}
\hfill
\subfigure[Reduction Cell]{\includegraphics[width=0.49\linewidth]{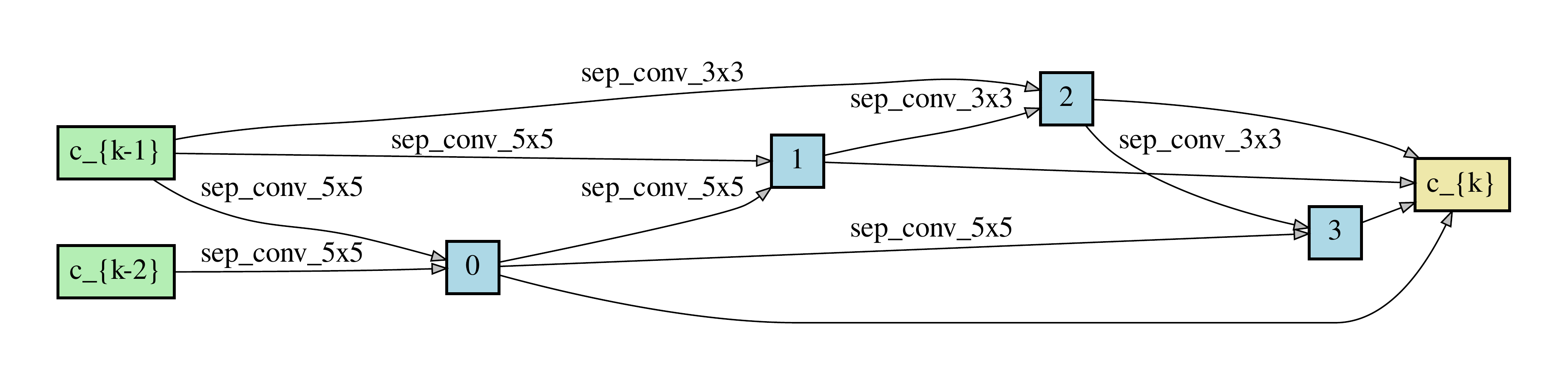}}
\caption{Normal and Reduction cells discovered by FP-NAS + TEG-NAS on DARTS space on CIFAR-10.}
\label{fig:ProbNAS_TEG_cifar10}
\end{figure}

We visualize the searched normal and reduction cells on DARTS space, by Reinforcement Learning (Figure \ref{fig:RL_TEG_cifar10}), Evolution (Figure \ref{fig:Evolution_TEG_cifar10}), and FP-NAS (Figure \ref{fig:ProbNAS_TEG_cifar10})).

\subsection{Ablation Study on $\hat{\kappa}$, $\hat{R}$, and $\mathrm{MSE}$}

To validate the necessity of considering all of the trainability, expressivity, and generalization, we conduct an ablation study in Table \ref{table:ablation} using Reinforcement Learning on Cifar100 on NAS-Bench-201.
This ablation study is conducted under the same settings as in Section~\ref{sec:201}.
As the baseline method, the RL agent uses the test accuracy after 1-epoch training as the reward. We can see that the guidance from every single indicator outperforms the truncated training, with much less search time cost. Finally, we achieve the best search performance once equipped with all $\hat{\kappa}$, $\hat{R}$, and $\mathrm{MSE}$.

\section{Conclusion}
We proposed a \textbf{unified} and \textbf{visualizable} NAS framework that benefit both various popular search methods and search interpretation.
We successfully disentangle the network's characteristics into three distinct aspects: \textit{\textbf{T}rainability, \textbf{E}xpressivity, \textbf{G}eneralization}, or ``\textbf{TEG}'' for short, and leverage all of them to provide effective and efficient guidance for NAS search.
Extensive studies on different NAS search methods validate the superior performance of our TEG-NAS framework. More importantly, we for the first time visualize the search trajectory on architecture landscapes from different search spaces, contributing to a better undertanding of both search and geometry of architecture space.
We hope our work encourages the community to further explore NAS methods that benefit from extremely low cost, and provide a better understanding of the architectures and complexity of different search spaces.

{\small
  \bibliographystyle{IEEEtran}
  \bibliography{conferences_abrv,references}
}

\begin{IEEEbiography}[{\includegraphics[width=1in,height=1.25in,clip,keepaspectratio]{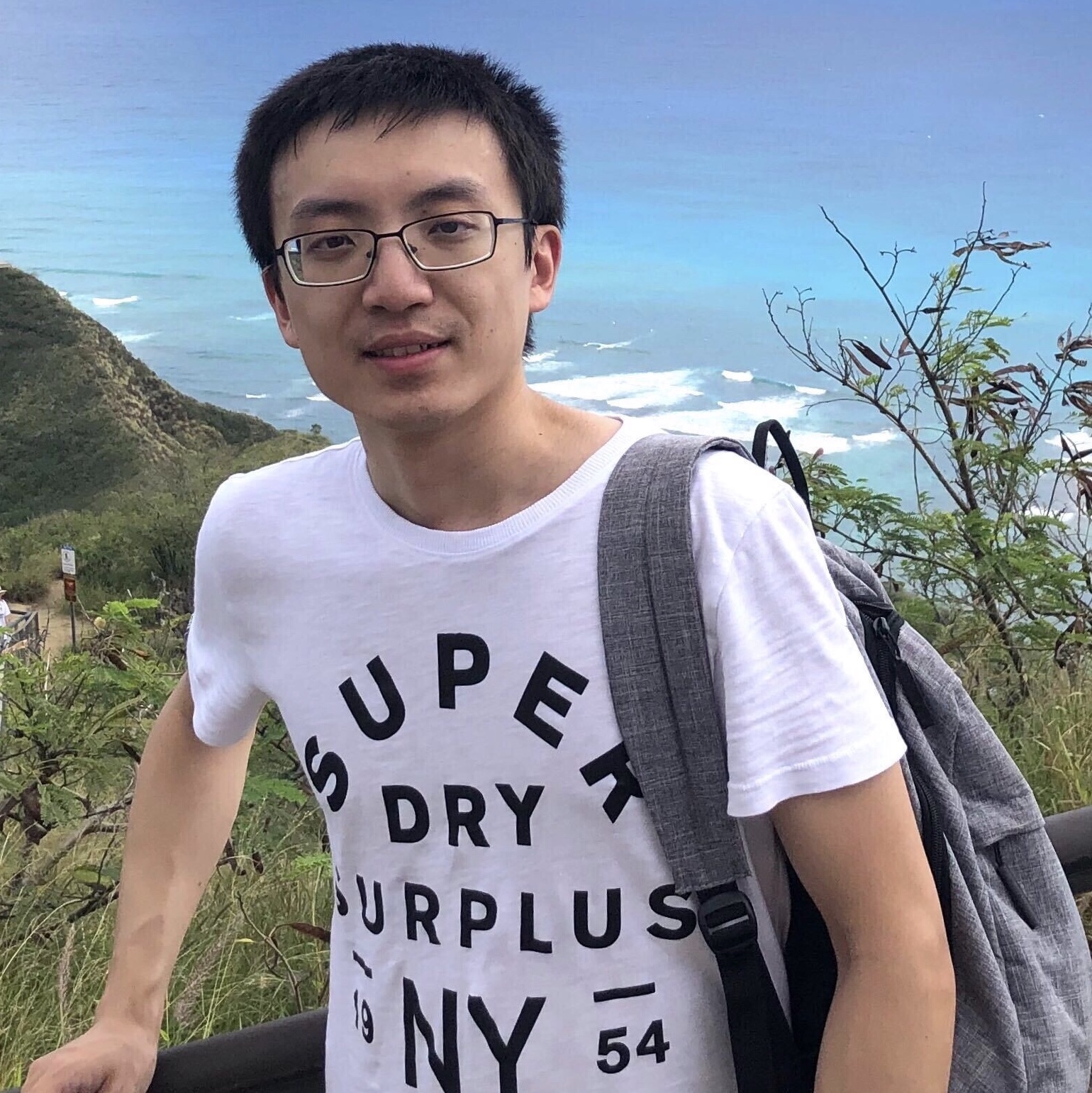}}]{Wuyang Chen}
is a Ph.D. student in Electrical and Computer Engineering at University of Texas at Austin. He received his M.S. degree in Computer Science from Rice University in 2016, and his B.S. degree from University of Science and Technology of China in 2014. His research focuses on addressing domain adaptation/generalization, self-supervised learning, and AutoML.
\end{IEEEbiography}

\begin{IEEEbiography}[{\includegraphics[width=1in,height=1.25in,clip,keepaspectratio]{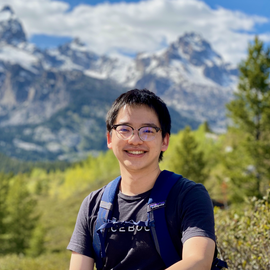}}]{Xinyu Gong}
is a Ph.D. student at the Department of Electrical and Computer Engineering, University of Texas at Austin. Before that, he obtained his Bachelor degree in Computer Science from the University of Electronic Science and Technology of China. He has also interned at Facebook AI, Horizon Robotics and Tencent AI Lab. His research interests are broadly in computer vision and machine learning, with a recent focus on neural architecture search.
\end{IEEEbiography}

\begin{IEEEbiography}[{\includegraphics[width=1in,height=1.25in,clip,keepaspectratio]{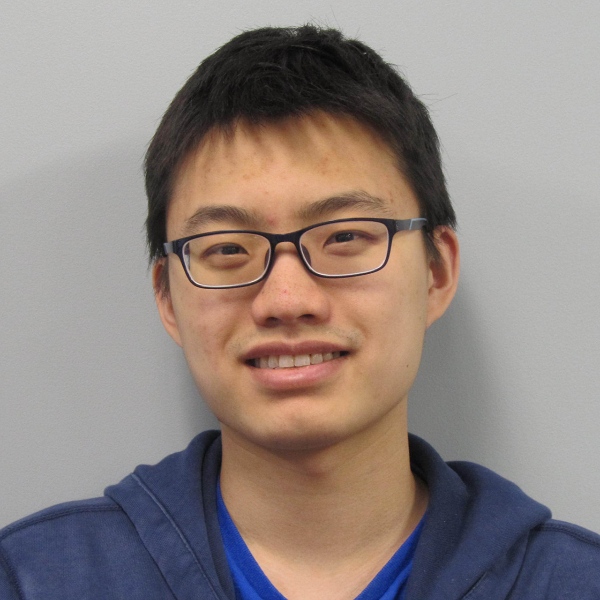}}]{Junru Wu}
is a Ph.D. student at the Department of Computer Science and Engineering at Texas A\&M University. Before that, he obtained his B.S. degree from Tongji University. He's interned at industry research labs including Google Research, Microsoft Research, NEC Labs America, and ByteDance AI Lab. His research interests lie in the intersection of computer vision and machine learning. In particular, He is interested in enabling efficient machine learning in a broad spectrum of computer vision problems, which includes Low-level Vision, Neural Architecture Search, and Multimodal Understanding.
\end{IEEEbiography}

\begin{IEEEbiography}[{\includegraphics[width=1in,height=1.25in,clip,keepaspectratio]{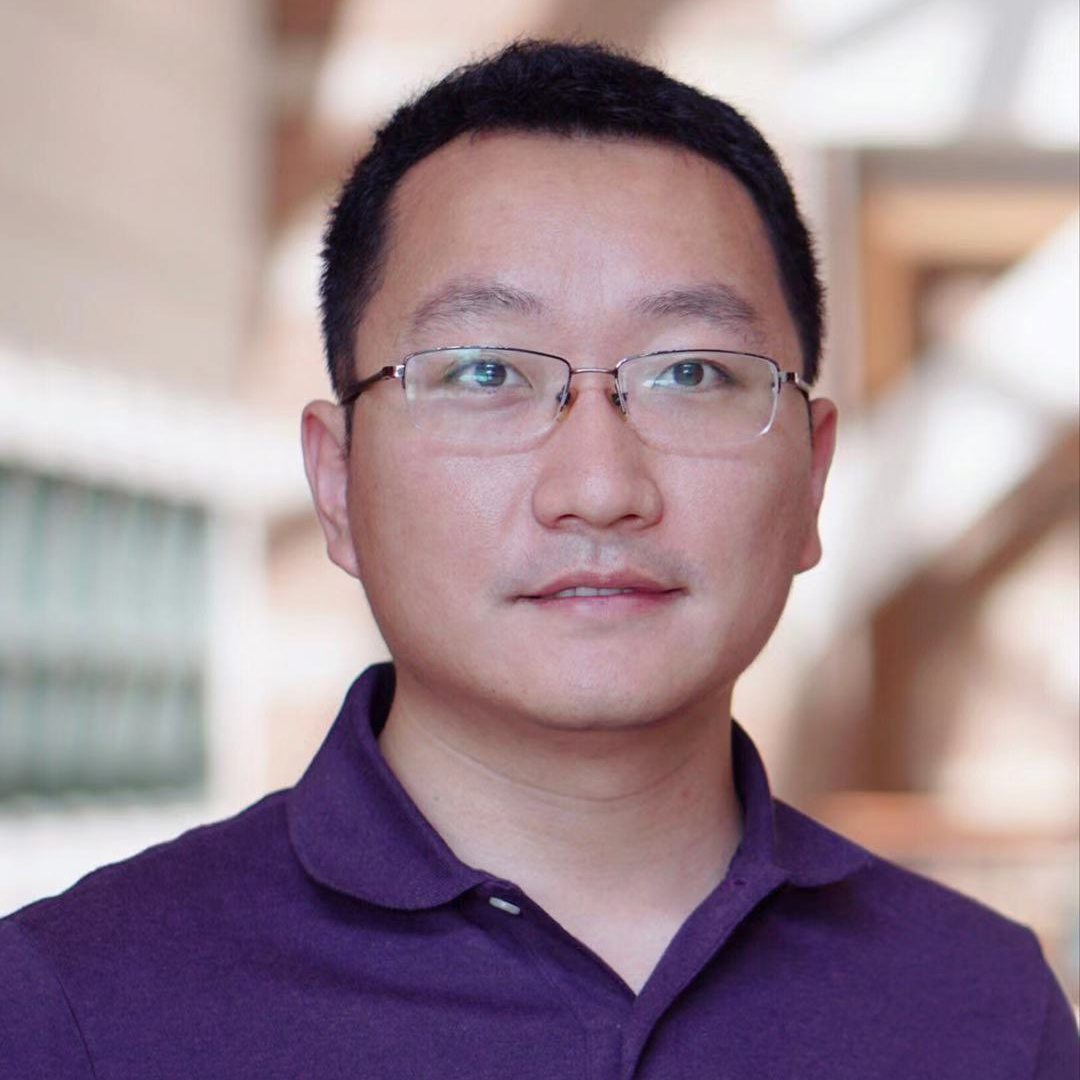}}]{Yunchao Wei} is currently a professor at the Center of Digital Media Information Processing, Institute of Information Science, at Beijing Jiaotong University. He received his Ph.D. degree from Beijing Jiaotong University, Beijing, China, in 2016. He was a Postdoctoral Researcher at Beckman Institute, UIUC, from 2017 to 2019. He is ARC Discovery Early Career Researcher Award Fellow from 2019 to 2021. His current research interests include computer vision and machine learning.
\end{IEEEbiography}

\begin{IEEEbiography}[{\includegraphics[width=1in,height=1.25in,clip,keepaspectratio]{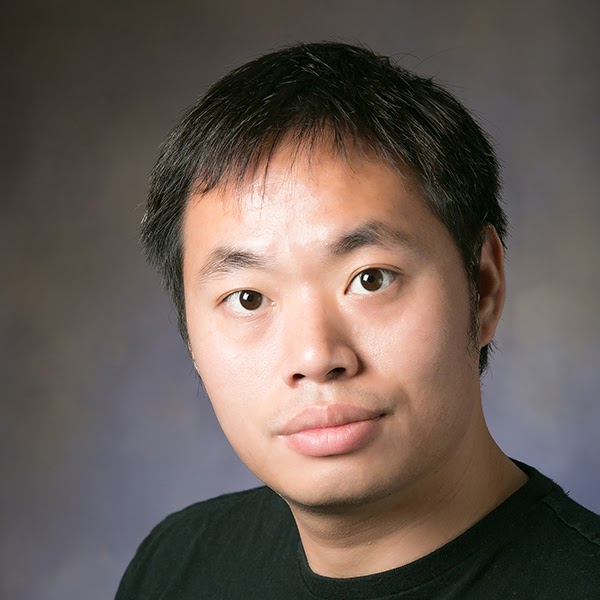}}]{Humphrey Shi} is an assistant professor in Department of Computer \& Information Science, University of Oregon. Dr. Shi joins the UO from IBM’s TJ Watson Research Center and the University of Illinois Grainger College of Engineering. His current research is at the intersection of computer vision and machine learning. Dr. Shi received awards from the Intelligence Advanced Research Projects Activity (IARPA) and winning multiple major international AI competitions such as the ImageNet video recognition challenge and the Nvidia AI City Challenge.
\end{IEEEbiography}

\begin{IEEEbiography}[{\includegraphics[trim={0 20cm 0 0},width=1in,height=1.25in,clip,keepaspectratio]{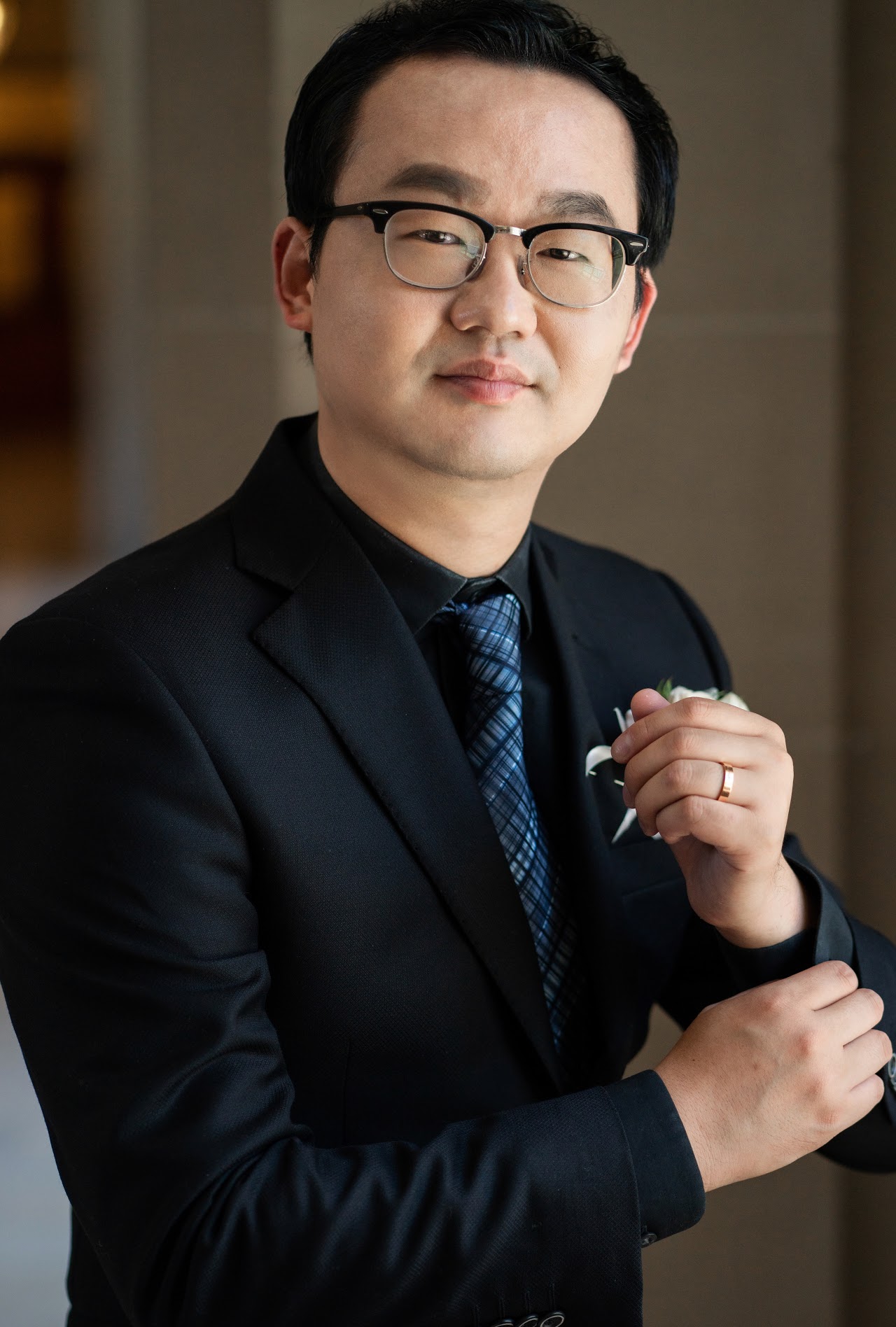}}]{Zhicheng Yan} Zhicheng Yan is a Staff Research Scientist at Facebook Research. He leads a team to develop a deep and personalized understanding of the objects and environment in the real world for improving Facebook AR products. Previsouly, he also worked on large-scale image and video understanding platforms. His research interests mainly include computer vision and machine learning. 
Zhicheng received his Ph.D from Department of Computer Science, University of Illinois at Urbana-Champaign in 2016. His supervisor is Prof. Yizhou Yu. Before that, he was a master student in the College of Computer Science and Technology, Zhejiang University and a research assistant of State Key Lab of CAD\&CG. His supervisor is Prof. Wei Chen. Zhicheng completed his Bachelor's degree at Zhejiang University majoring in Software Engineering in July, 2007.
\end{IEEEbiography}
\begin{IEEEbiography}[{\includegraphics[trim={0 1.5cm 0 0.2cm},width=1in,height=1.25in,clip,keepaspectratio]{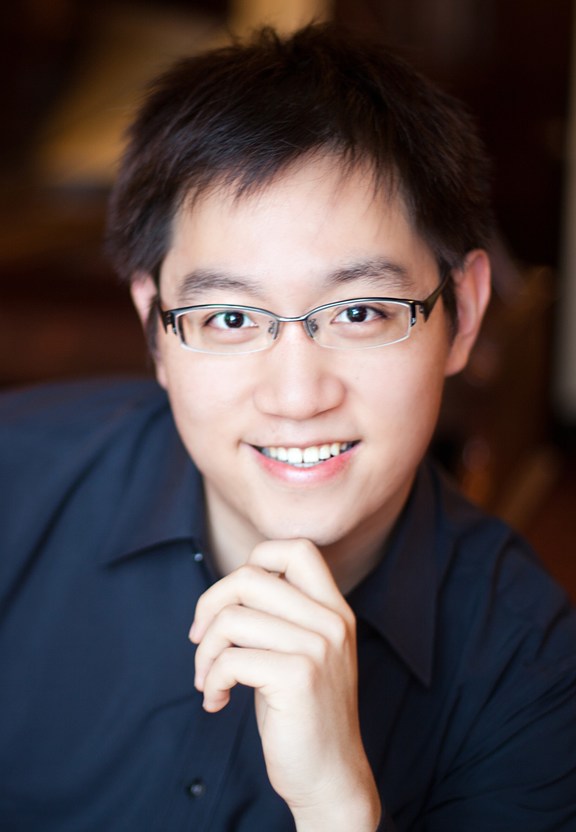}}]{Yi Yang} is currently a Professor at the Zhejiang University. He received his Ph.D. degree from Zhejiang University, Hangzhou, China, in 2010. He was a post-doctoral researcher in the School of Computer Science, Carnegie Mellon University. His current research interests include machine learning and its applications to multimedia content analysis and computer vision, such as multimedia retrieval and video content understanding.
\end{IEEEbiography}
\begin{IEEEbiography}[{\includegraphics[trim={0 10cm 0 3cm},width=1in,height=1.25in,clip,keepaspectratio]{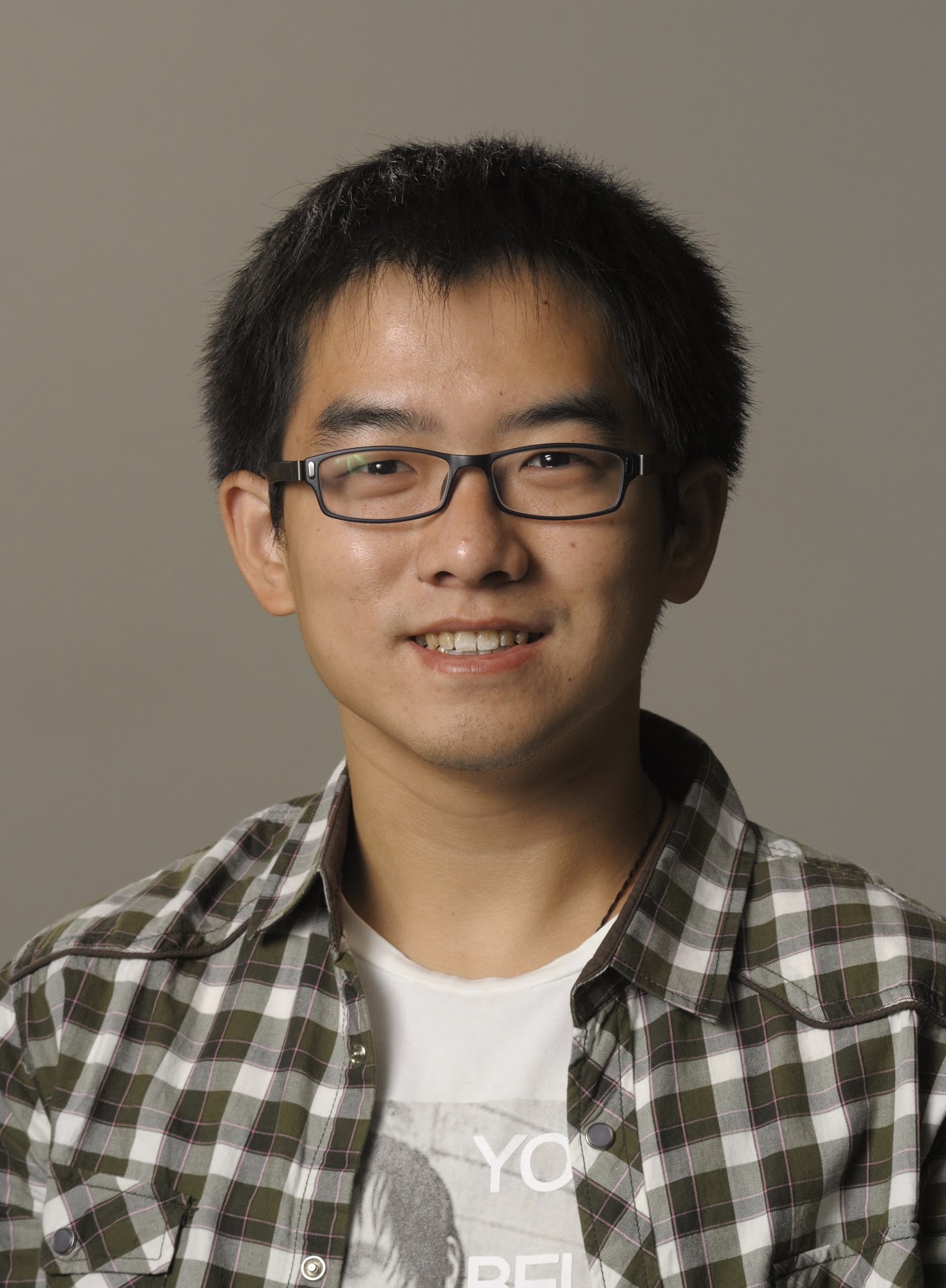}}]{Zhangyang Wang}
is currently an Assistant Professor of ECE at UT Austin. He was an Assistant Professor of CSE, at TAMU, from 2017 to 2020. He received his Ph.D. in ECE from UIUC in 2016, and his B.E. in EEIS from USTC in 2012. Prof. Wang is broadly interested in the fields of machine learning, computer vision, optimization, and their interdisciplinary applications. His latest interests focus on automated machine learning (AutoML), learning-based optimization, machine learning robustness, and efficient deep learning.
\end{IEEEbiography}

\end{document}